\documentclass[letterpaper]{article} 
\usepackage{aaai2027}  
\usepackage[hyphens]{url}  
\usepackage{graphicx} 
\usepackage{natbib}  
\usepackage{caption} 
\usepackage{algorithm}
\usepackage{algorithmic}
\usepackage{tabularx}
\usepackage{amsmath}
\usepackage{colortbl}
\usepackage{xcolor}
\usepackage{multirow}
\usepackage[most]{tcolorbox}
\usepackage{placeins}
\usepackage{makecell}
\usepackage{longtable}

\usepackage{newfloat}
\usepackage{listings}
\DeclareCaptionStyle{ruled}{labelfont=normalfont,labelsep=colon,strut=off} 
\floatstyle{ruled}
\newfloat{listing}{tb}{lst}{}
\floatname{listing}{Listing}

\usepackage{booktabs}

\newcommand{\ours}{TableSeek}
\title{\ours{}: Structure-Preserving Agentic Evidence Seeking \\ over Heterogeneous Table Corpora}
\author{
    Jiaming Tian\textsuperscript{\rm 1},
    Liyao Li\textsuperscript{\rm 1},
    Wentao Ye\textsuperscript{\rm 1},
    Haobo Wang\textsuperscript{\rm 1},\\
    Lihua Yu\textsuperscript{\rm 2},
    Zujie Ren\textsuperscript{\rm 1,\rm 3}\corresponding,
    Gang Chen\textsuperscript{\rm 1},
    Junbo Zhao\textsuperscript{\rm 1}
}
\affiliations{
    \textsuperscript{\rm 1}Zhejiang University\\
    \textsuperscript{\rm 2}Bank of Hangzhou Co., Ltd.\\
    \textsuperscript{\rm 3}Zhejiang Lab\\
    hsxz2@zju.edu.cn,
    renzju@zju.edu.cn
}

\begin{document}

\maketitle

\begin{abstract}
Open-domain table retrieval seeks tables that contain sufficient evidence for answering a question or verifying a claim. Yet semantic relevance is often misleading: topically similar tables may lack the required facts, while answer-bearing evidence is often confined to a few cells whose meaning depends on surrounding schema and table context. Heterogeneous schemas, value formats, and serializations further weaken one-shot matching.

We present \textbf{\ours{}}, a structure-preserving agentic search framework for heterogeneous table corpora. Instead of ranking tables once, an LLM agent iteratively follows sparse clues, inspects schema-preserving previews, identifies schema- and value-level mismatches, and refines its investigation. \ours{} uses cells and schemas as evidence anchors while retaining complete tables as evidence units, enabling fine-grained localization without losing the context required for interpretation and answerability checking.

Without relying on retriever training or a precomputed semantic index, \ours{} produces transparent evidence-seeking trajectories and achieves competitive end-to-end performance against strong retrieval-and-reranking pipelines on heterogeneous table benchmarks. These results suggest that active, structure-preserving evidence seeking is a promising paradigm for open-domain table retrieval.
\end{abstract}

\section{Introduction}
\label{sec:introduction}

Tables are increasingly used as external knowledge sources for question answering, retrieval-augmented generation, and structured-data intelligence systems~\citep{WikiTQ,HybridQA,OTTQA,TabFact,TAPEX,Structgpt}. In these settings, systems must first retrieve tables that contain the evidence required to answer a question or verify a claim. Consequently, table retrieval has become a fundamental component of many table-centric intelligence pipelines~\citep{TARGET, JAR}.
Recent work shows that \textbf{semantic relevance can diverge substantially from answerability and evidence sufficiency} in open-domain table retrieval~\citep{TCR-Bench}. A retrieved table may discuss the correct topic while lacking the facts needed to support an answer. Conversely, evidence-supporting tables may be overlooked because relevant information is expressed through different schemas, terminology, or value representations. This challenge is particularly pronounced in heterogeneous table corpora, where equivalent information can appear under diverse schemas and serializations.

\begin{figure}[t]
    \centering
    \includegraphics[width=0.992\linewidth]{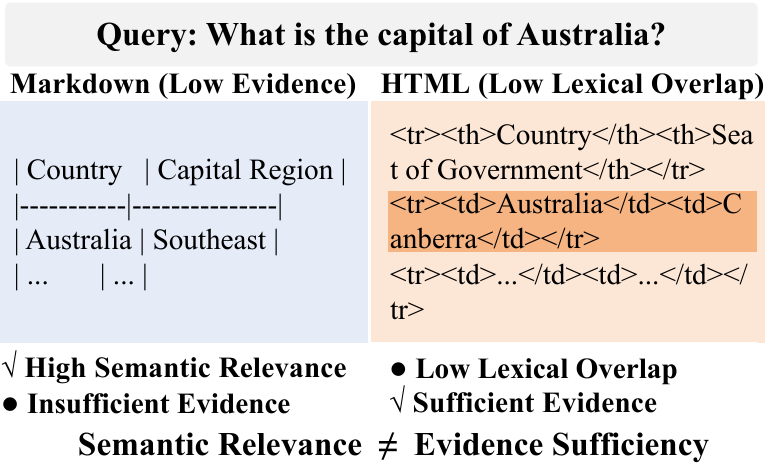}
    \caption{Semantic relevance $\neq$ evidence sufficiency. The left table mentions \texttt{Australia} and \texttt{Capital Region} but lacks the actual capital city. The right table contains the exact answer \texttt{Canberra}, but uses \texttt{Seat of Government} instead of \texttt{Capital}, resulting in low lexical overlap with the query.}
    \label{fig:motivation}
\end{figure}

Existing retrieval systems primarily address this challenge through increasingly sophisticated ranking mechanisms. Sparse retrievers rely on lexical matching, while dense retrievers and rerankers estimate semantic relevance using learned representations~\citep{BM25,DPR,Contriever,Colbert,NQ-TABLES-DTR,THYME}. Despite their differences, these approaches largely treat retrieval as a one-shot ranking problem, assuming that evidence-supporting tables can be identified directly from relevance estimates. However, evidence sufficiency cannot always be inferred from relevance alone. In heterogeneous repositories, relevant evidence is often expressed through different schemas, abbreviations, naming conventions, or value formats. As a result, retrieval failures may arise not because evidence is absent, but because the retrieval process cannot bridge these representation mismatches.

To address this challenge, we propose \textbf{\ours{}}, a structure-preserving agentic retrieval framework for heterogeneous table corpora. Rather than relying on a single relevance estimate, \ours{} formulates retrieval as an \textbf{iterative evidence-seeking process}. Starting from an initial search hypothesis, an LLM agent searches for sparse clues, inspects retrieved tables, identifies schema- and value-level mismatches, and refines subsequent search strategies. Retrieval decisions are therefore guided by accumulated evidence rather than the original query alone.

A key observation underlying our design is that table evidence exhibits dual granularity. Useful clues are often localized to a few cells, making fine-grained search desirable, while interpreting those clues requires schema information, row-column relationships, and table-level context. Motivated by this observation, \ours{} uses cells and schemas as search anchors while retaining complete tables as evidence units, enabling fine-grained evidence localization without sacrificing structural context.

This formulation is particularly well suited to heterogeneous table retrieval. Retrieval failures themselves often reveal schema conventions, naming patterns, and value representations that can guide subsequent search. Consequently, retrieval can proceed as an iterative interaction between an LLM agent and the repository, progressively refining retrieval hypotheses based on observed evidence. Search effort can also adapt naturally to query difficulty, allocating additional exploration only when necessary~\citep{DCI,GrepRAG}.
By directly interacting with raw table contents and adapting search behavior through evidence-driven feedback, \ours{} accommodates schema heterogeneity without requiring retriever training, dense indexing, reranking models, or offline semantic indexes.


\textbf{Our contributions are threefold.} \textbf{(1) Problem formulation:} motivated by recent findings on evidence sufficiency, we formulate heterogeneous table retrieval as an iterative evidence-seeking process rather than a one-shot relevance ranking problem. \textbf{(2) Method:} we propose \textbf{\ours{}}, a structure-preserving agentic retrieval framework that combines fine-grained evidence search, schema-preserving table inspection, and gap-guided iterative refinement, using cells and schemas as search anchors while retaining complete tables as evidence units. \textbf{(3) Empirical evidence:} we demonstrate that \ours{} consistently improves evidence-oriented table retrieval across heterogeneous benchmarks and achieves competitive end-to-end QA performance without retriever training, dense indexing, or reranking models. Code is publicly available at \url{https://github.com/minger-hsxz/TableSeek-Open}.

\section{Related Work}

\subsection{Agentic Retrieval}

Recent work has explored retrieval systems that move beyond static top-$k$ ranking and allow language models to interact directly with external repositories. ReAct~\citep{REACT} and Search-o1~\citep{Search-o1} integrate retrieval actions into the reasoning process, while Self-RAG~\citep{Self-RAG} and broader agentic RAG systems incorporate retrieval-aware reflection, planning, tool use, and iterative refinement~\citep{Agentic_RAG-survey}. Beyond retrieval-augmented generation, recent agent frameworks have investigated evidence tracking, working memory, and context engineering for multi-step evidence gathering and reasoning~\citep{TRACK-RANK-CRACK,ACE}. DCI~\citep{DCI} formulates retrieval as direct corpus interaction through executable search operations, and GrepRAG~\citep{GrepRAG} demonstrates the effectiveness of grep-style retrieval combined with iterative refinement.

These methods are primarily designed for text or code repositories. In contrast, tabular evidence depends heavily on schema semantics and row-column structure, and retrieval failures often arise from representation mismatches rather than missing information. \ours{} extends agentic retrieval to heterogeneous table repositories through structure-preserving evidence seeking and schema-guided refinement.

\subsection{Table Retrieval}

Most table retrieval methods follow a retrieval-as-ranking paradigm. Prior work adapts dense retrieval to tables~\citep{NQ-TABLES-DTR,DPR}, develops table-aware representations and hybrid matching strategies~\citep{table-specific-embedding-not-better,TEM,THoRR,THYME}, and designs efficient multi-stage retrieval pipelines~\citep{CRAFT}. Other studies explore more challenging settings such as join-aware retrieval across related tables~\citep{JAR}. Meanwhile, benchmarks including NQ-Tables~\citep{NQ-TABLES-DTR}, TARGET~\citep{TARGET}, JAR~\citep{JAR}, and TCR-Bench~\citep{TCR-Bench} have highlighted the difficulty of retrieving evidence-supporting tables from large and heterogeneous collections.

Unlike prior work that primarily improves table representation, matching, or ranking, \ours{} focuses on the retrieval process itself. Rather than relying on a single ranking stage, the framework iteratively searches for evidence, diagnoses retrieval failures from retrieved tables, and refines subsequent search strategies. Our work therefore bridges recent advances in agentic retrieval and table intelligence, bringing iterative evidence seeking to open-domain table retrieval.

\begin{figure*}[t]
\centering
\includegraphics[width=0.95\linewidth]{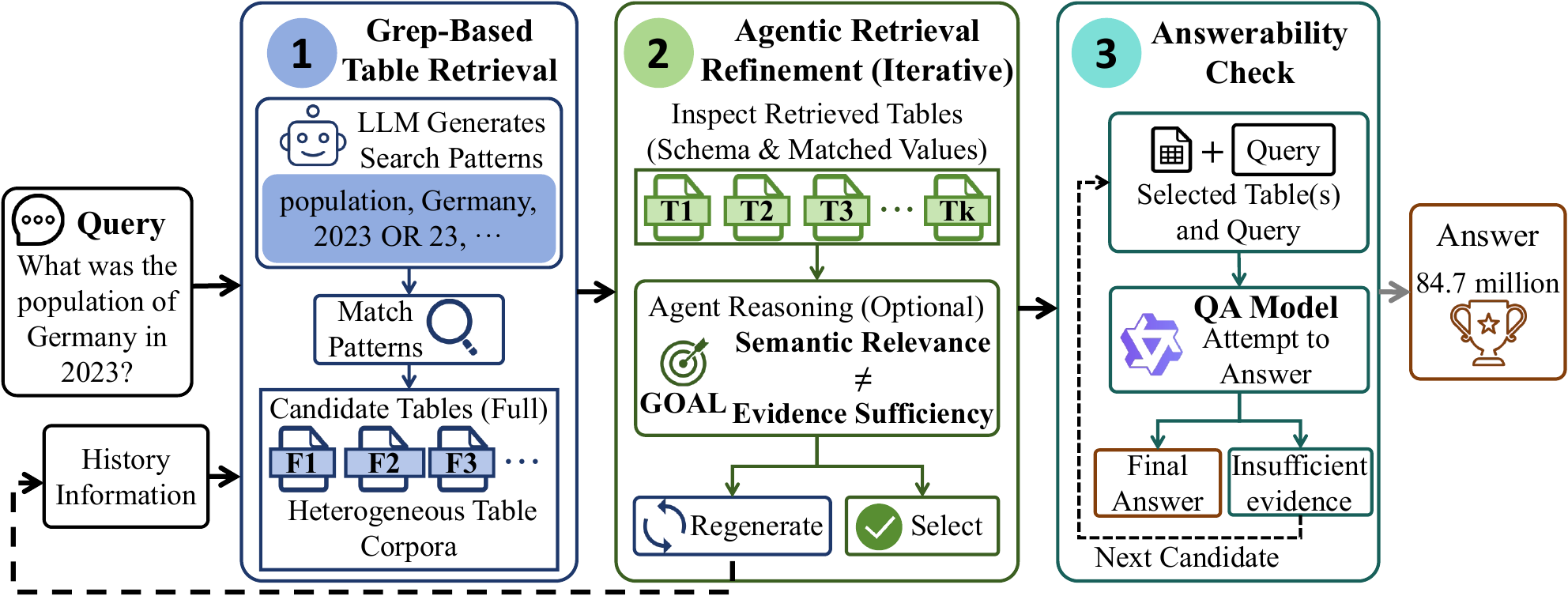}
\caption{Overview of \ours{} for question answering over heterogeneous table corpora. Stage 1 scores tables against generated search patterns to form a ranked candidate list. Stage 2 inspects table previews and iteratively refines the patterns until a candidate is selected. Stage 3 checks answerability on the selected table, falling back to the next candidate when evidence is insufficient. Here, $F_i$ denotes a table format, and $T_i$ denotes the $i$-th ranked table.}
\label{fig:framework}
\end{figure*}

\section{Method}
\label{sec:method}

\subsection{Overview}

We study open-domain question answering over heterogeneous table repositories. Given a query $q$ and a table collection $\mathcal{T}=\{T_1,T_2,\ldots,T_N\}$, the goal is to identify tables containing sufficient evidence for answering the query.

We formulate retrieval as an iterative interaction between an LLM agent and the table repository. Rather than relying on a single retrieval step, the agent searches for evidence, inspects retrieved tables, identifies missing information, and refines subsequent search strategies.

As shown in Figure~\ref{fig:framework}, \ours{} consists of three stages: (1) grep-based table retrieval, (2) agentic retrieval refinement, and (3) answerability check.

\subsection{Grep-Based Table Retrieval}

Given a query $q$, an LLM first generates a set of grep-style search patterns $\mathcal P=\{p_1,p_2,\ldots,p_m\}$ covering entities, attributes, values, and schema concepts that may appear in evidence-supporting tables.

Unlike passage retrieval, \ours{} treats complete tables as evidence units. For each table $T_i$, we compute

\begin{equation}
Score(T_i)=\frac{1}{|\mathcal P|}
\sum_{p\in\mathcal P}
\mathbf 1[p\in T_i].
\end{equation}

Tables are ranked by matching scores to obtain the top-$k$ candidates $\mathcal R=(T_{(1)},T_{(2)},\ldots,T_{(k)})$.

\subsection{Agentic Retrieval Refinement}

A single retrieval round may miss relevant tables due to schema discrepancies, abbreviations, or value variations. To address this, the agent examines the top-$k$ retrieved candidates, which are presented in a \textbf{structure-preserving} manner, either as full tables when small or as serialized headers with a few matched rows when large, and then decides between \texttt{select} and \texttt{regenerate}.

The \texttt{select} operation further prunes the top-$k$ retrieved candidate tables, retaining a subset for downstream verification, while the \texttt{regenerate} operation refines the search patterns based on the selected subset and initiates another round of retrieval. The refinement process can be summarized as \( \mathcal P_t \rightarrow \mathcal R_t \rightarrow \mathcal P_{t+1} \), where \( \mathcal P_t \) denotes the current pattern set, \( \mathcal R_t \) the top-$k$ retrieval results, and \( \mathcal P_{t+1} \) the updated patterns for the next iteration.

This allows retrieval decisions to be conditioned on previously retrieved evidence and its effectiveness, rather than solely on the original query.

\subsection{Answerability Check}

Retrieved tables may still lack sufficient evidence even when they appear relevant. After a table is selected, the QA model attempts to answer the query using the table content. If the response indicates insufficient evidence, the table is rejected, and the system proceeds with the \texttt{fallback} procedure. That is, it discards the current table, retrieves the next candidate from the ranked list, and re-attempts QA on the new table. This fallback procedure repeats until either a satisfactory answer is produced or the exploration budget is exhausted. If the budget is depleted during the refinement stage, the system terminates the refinement process and directly performs QA on the current top-1 retrieved table.

Together, retrieval refinement and answerability checking form an iterative evidence-seeking process, where retrieval decisions are continuously updated based on observed evidence rather than query similarity alone. This design enables \ours{} to better handle heterogeneous table structures and representation variations.

\section{Experiments}

\subsection{Experimental Setup}

\subsubsection{Benchmarks}
We primarily evaluate on two benchmark families, summarized in Table~\ref{tab:bench_stats}.

\textbf{TCR-Bench~\citep{TCR-Bench}.}
TCR-Bench serves as our primary benchmark. It provides a heterogeneous table corpus with large tables, hard negatives that are semantically related to the query yet do not contain the answer, and semantic augmentation. We evaluate all four formats (Markdown, CSV, HTML, Mixed). Mixed is a predefined mixture of the other three. We adopt Mixed as our primary setting.

\textbf{TARGET-TabFact~\citep{TARGET}.}
We use the TabFact~\citep{TabFact} subset of TARGET as a complementary benchmark. Unlike TCR-Bench, it is derived from a table verification task in which queries correspond to claims about tables. Supporting claims evaluate evidence retrieval, while non-supporting claims test robustness under strong semantic overlap.

\begin{table}[t]
\centering
\small
\begin{tabular}{llrr}
\toprule
Dataset & Split & \#Queries & \#Tables \\
\midrule
TCR-Bench & -- & 209 & 637 \\
TARGET-TabFact & valid & 12,792 & 1,696 \\
TARGET-TabFact & valid-1k & 1,000 & 1,696 \\
TARGET-TabFact & test & 12,779 & 1,695 \\
TARGET-TabFact & test-1k & 1,000 & 1,695 \\
\bottomrule
\end{tabular}
\caption{Statistics of benchmarks and table corpora used in our experiments.}
\label{tab:bench_stats}
\end{table}

\subsubsection{Models and Baselines}
We compare against dense retrieval and reranker-based baselines. Our primary retrieval baselines use Qwen3-Embedding-4B (E4B)~\citep{qwen3embedding} and \texttt{stella\_en\_1.5B\_v5} (Stella)~\citep{Stella}, while Qwen3-Reranker-4B (R4B) is used to rerank the top-$k$ retrieved candidates, a setup we denote as \textbf{Rerank (top-$k$)}.

For \ours{}, we primarily evaluate three backbone LLMs: Qwen3-30B-A3B-Thinking-2507 (Qwen3-30B), Qwen3-4B-Instruct-2507 (Qwen3-4B) \citep{qwen3technicalreport}, and Llama-3.1-8B-Instruct (LLaMA) \citep{LLaMA}. Qwen3-30B serves as the main backbone, offering a balance between reasoning capability and inference efficiency for iterative retrieval refinement, while Qwen3-4B provides a lightweight configuration. LLaMA is included to evaluate \ours{} across different model families. Additional model configurations, variants, and implementations are detailed in the appendix.

\subsubsection{Main Evaluation Metrics}
We adopt two sets of metrics.

\textbf{Retrieval.}
For dense and reranked retrieval, we report Recall@$k$ (R@$k$)~\citep{MultiTableQA, t2-ragbench}, i.e., whether the gold table is in top-$k$ results. For \ours{}, we report the recall of the gold table across all QA calls (R@all).

\textbf{End-to-end QA.}
On TCR-Bench, we report F1. On TARGET-TabFact, we report overall accuracy (Acc) together with support accuracy (Acc-Sup) and refute accuracy (Acc-Ref). Since retrieval quality primarily affects support examples that require evidence-bearing tables, we mainly focus on Acc and Acc-Sup. Following the benchmark objectives, we regard end-to-end QA performance as the primary metric, since it directly measures whether retrieved evidence is sufficient for solving the downstream task.

\subsubsection{Implementation Details}
All experiments are conducted using local deployment on NVIDIA GPUs. Embedding, reranking, and LLM inference are implemented with \texttt{transformers} and \texttt{sglang} (deterministic inference). Unless otherwise noted, each algorithm is evaluated with a single run. Additional deployment details are provided in the appendix.

\subsection{Main Results}
\label{sec:main_results}

Tables~\ref{tab:main_tcr} and \ref{tab:main_tabfact} compare dense retrieval, reranking, and \ours{} on TCR-Bench (Mixed format) and TARGET-TabFact-test-1k, respectively, using Qwen3-30B as the backbone. For each benchmark, the reranking results are obtained by applying the reranker on top of the best-performing embedding model's retrieval outputs. Results for additional backbones (Qwen3-4B, LLaMA), other table formats, and oracle upper-bound settings are provided in the appendix.

\begin{table}[t]
\centering
\small
\begin{tabular}{llcc}
\toprule
Method & Config & Recall & F1 \\
\midrule
Dense & E4B & 0.172 & 0.118 \\
\midrule
Rerank (top-3) & E4B+R4B & 0.359 & 0.269 \\
Rerank (top-5) & E4B+R4B & 0.421 & 0.317 \\
\midrule
\ours{} (5-step) & Qwen3-30B & 0.416 & 0.284 \\
\ours{} (10-step) & Qwen3-30B & \textbf{0.507} & \textbf{0.361} \\
\bottomrule
\end{tabular}
\caption{Main results on TCR-Bench (Mixed format) with Qwen3-30B as the QA/backbone model. Recall is R@1 for Dense/Rerank and R@all for \ours{}. Full results with additional backbones are given in the appendix.}
\label{tab:main_tcr}
\end{table}

\begin{table}[t]
\centering
\setlength{\tabcolsep}{1.3mm} 
\small
\begin{tabular}{llcccc}
\toprule
Method & Config & Recall & Acc & Acc-Sup \\
\midrule
Dense & Stella & 0.519 & 0.721 & 0.489 \\
\midrule
Rerank (top-3) & Stella+R4B & 0.623 & 0.752 & 0.562 \\
Rerank (top-5) & Stella+R4B & \textbf{0.656} & 0.754 & 0.572 \\
\midrule
\ours{} (5-step) & Qwen3-30B & 0.617 & 0.754 & 0.566 \\
\ours{} (10-step) & Qwen3-30B & 0.648 & \textbf{0.764} & \textbf{0.589} \\
\bottomrule
\end{tabular}
\caption{Results on TARGET-TabFact-test-1k with Qwen3-30B as the QA/backbone model. Retrieval is R@1 for Dense/Rerank and R@all for \ours{}.}
\label{tab:main_tabfact}
\end{table}

On TCR-Bench, \ours{} achieves performance comparable to strong retrieval-and-reranking pipelines despite requiring neither retriever training nor a dedicated reranker. Under a 10-step budget, \ours{} attains the highest retrieval score (\textbf{0.507}) and F1 (\textbf{0.361}), compared to 0.421 retrieval and 0.317 F1 for the strongest retrieval-based baseline, Rerank (top-5).

On TARGET-TabFact-test-1k, \ours{} (10-step) achieves the highest overall accuracy (\textbf{0.764}) and supported-case accuracy (\textbf{0.589}) among all compared methods.

Overall, results across both benchmarks demonstrate the effectiveness of iterative, structure-preserving evidence seeking in heterogeneous table retrieval. Additional results on other baseline models, alternative input formats, and statistical significance tests are provided in the appendix.

\subsection{Discriminative Score: Semantic Relevance vs. Evidence Sufficiency}
Following TCR-Bench, Top-$k$ Group Recall ($GR@k$) measures whether retrieval reaches the correct topical neighborhood by retrieving any table from the query's Sibling Table Group. The Discriminative Score, defined as $DS@k = R@k / GR@k$, evaluates how effectively a method identifies the gold table once that neighborhood has been reached. Intuitively, lower DS values indicate a stronger emphasis on semantic relatedness, whereas higher values reflect a greater tendency to distinguish the evidence-bearing table from semantically similar alternatives.

For \ours{}, we report $DS@all$, measuring evidence discrimination across the entire search trajectory, and $DS@final$, evaluating the final evidence selection used for answer generation. As shown in Table~\ref{tab:ds_results}, \ours{} achieves substantially higher DS values than Rerank (top-5) ($DS@all=0.803$ and $DS@final=0.750$ for the 10-step setting, versus $DS@1=0.540$) This suggests that the two paradigms prioritize different retrieval signals: retrieval-and-reranking methods focus more on semantic relatedness, whereas \ours{} places greater emphasis on identifying tables that directly contain the required evidence.

\begin{table}[t]
\centering
\small
\begin{tabular}{lccccc}
\toprule
\multirow{2}{*}{Metric} & \multirow{2}{*}{Dense} & \multicolumn{2}{c}{Rerank} & \multicolumn{2}{c}{\ours{}} \\
\cmidrule(lr){3-4} \cmidrule(lr){5-6}
 &  & top-3 & top-5 & 5step & 10step \\
\midrule
R@1 & 0.172 & 0.359 & 0.421 & -- & -- \\
DS@1 & 0.252 & 0.478 & 0.540 & -- & -- \\
\midrule
R@all & -- & -- & -- & 0.416 & 0.507 \\
DS@all & -- & -- & -- & 0.725 & 0.803 \\
R@final & -- & -- & -- & 0.354 & 0.431 \\
DS@final & -- & -- & -- & 0.655 & 0.750 \\
\bottomrule
\end{tabular}
\caption{Recall and Discriminative Score on TCR-Bench. Single-pass baselines are evaluated at their single retrieval decision (R@1/DS@1); \ours{} is evaluated both across all QA calls (R@all/DS@all) and at the final call used for answer generation (R@final/DS@final).}
\label{tab:ds_results}
\end{table}

\subsection{Robustness to Heterogeneous Table Formats}
Real-world table corpora often contain multiple serialization formats. To evaluate robustness under format variation, we compare all methods across CSV, Mixed, HTML, and Markdown versions of TCR-Bench using Qwen3-30B throughout.

For each method, we report the mean F1 across formats together with the range, standard deviation, and Format Robustness Score (FRS). FRS is defined as the percentage of the best-format performance retained under the least favorable format: $\mathrm{FRS} = 100 \times \min(\mathrm{F1}) / \max(\mathrm{F1})$. A higher FRS indicates better robustness to format changes, with 100\% representing complete format invariance.


\begin{figure}[t]
\centering
\includegraphics[width=0.9\linewidth]{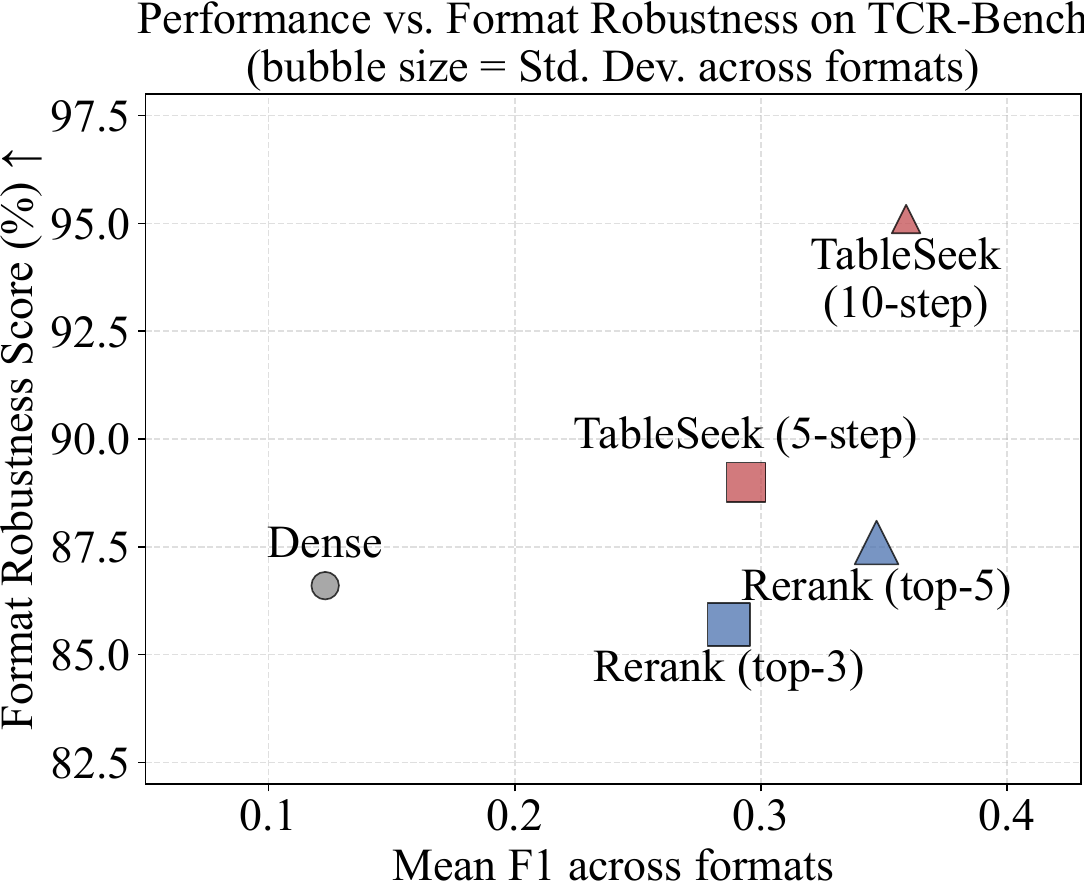}
\caption{F1 across table formats and robustness statistics on TCR-Bench, using Qwen3-30B.}
\label{fig:robustness}
\end{figure}

Figure~\ref{fig:robustness} summarizes the results.
Dense retrieval exhibits low variability across formats; however, this largely reflects uniformly low performance across all formats. Rerank methods achieve higher average F1 but show larger performance fluctuations under different serializations. In contrast, \ours{} (10-step) combines competitive performance (mean F1 = 0.359) with the highest robustness score (FRS = 95.1\%), retaining more than 95\% of its best-format performance under the least favorable format.

These results suggest that \ours{} is less sensitive to table serialization changes than conventional retrieval-and-reranking pipelines while maintaining strong overall performance. Additional robustness experiments with multiple runs and hyperparameter configurations are provided in the appendix and show consistent trends.

\subsection{Efficiency Analysis: API Calls and Runtime}

We analyze efficiency from two perspectives: inference call complexity and wall-clock runtime. Since wall-clock runtime depends on both computational workload and system implementation, we primarily focus on call complexity as a structural measure, while runtime results are reported as a complementary reference.

\textbf{Call complexity.} Let $N$ be the number of corpus tables, $M$ the number of queries, and $k$ the rerank depth. Dense retrieval requires one-time corpus encoding and per-query encoding plus one QA call per query, yielding $C_{\text{Dense}} = N + 2M$. Rerank adds $k$ reranking calls per query, giving $C_{\text{Rerank-}k} = N + (k+2)M$. Both methods thus incur a corpus-dependent preprocessing cost.

In contrast, \ours{} does not require corpus-level semantic indexing or learned retrieval models. For query $i$, let $s_i \le S$ be its agent steps. Total calls are $C_{\ours{}} = \sum_{i=1}^M s_i \le S M$, depending on query complexity, not corpus size. We set $S=5$ and $S=10$ to roughly match Rerank (top-3) and Rerank (top-5). As shown in Table~\ref{tab:calls_small}, Dense and Rerank have fixed counts, while \ours{} often stops before the budget, using fewer calls in practice.

\begin{figure}[t]
    \centering
    \includegraphics[width=0.95\columnwidth]{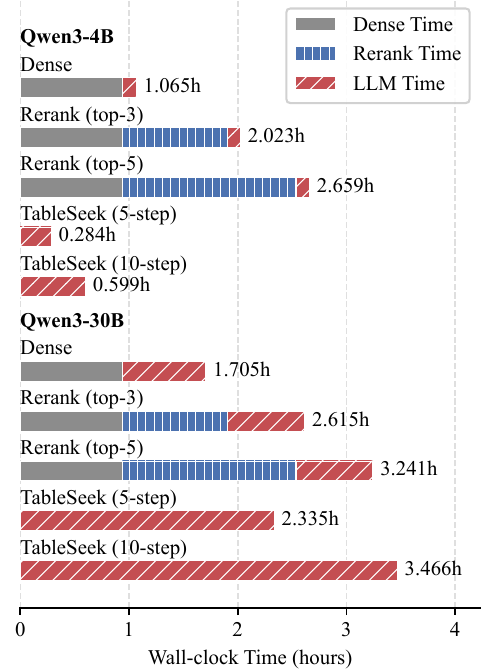}
    \caption{Wall-clock time breakdown by stage (dense encoding, reranking, LLM/QA) across methods on TCR-Bench.}
    \label{fig:time_breakdown}
\end{figure}

\textbf{Runtime.}
Figure~\ref{fig:time_breakdown} reports the wall-clock runtime on TCR-Bench. Since runtime depends on the dataset, implementation, and deployment environment, these results are intended as an empirical reference for this benchmark.

\begin{table}[t]
\centering
\small
\begin{tabular}{lcc}
\toprule
Method & Calls (4B) & Calls (30B) \\
\midrule
Dense      & 1055 & 1055 \\
Rerank (top-3)   & 1682 & 1682 \\
Rerank (top-5)   & 2100 & 2100 \\
\midrule
\ours{} (5-step)  & \textbf{918}  & \textbf{921} \\
\ours{} (10-step) & \textbf{1581} & \textbf{1305} \\
\bottomrule
\end{tabular}
\caption{Total inference calls on TCR-Bench. Dense/Rerank are independent of the QA backbone.}
\label{tab:calls_small}
\end{table}

\section{Ablation Studies}

\subsection{Effect of Refine, Fallback, and Structure Preservation}

We first ablate two core components of \ours{}: the \texttt{refine} mechanism, which enables iterative inspection and updating of retrieved evidence, and the \texttt{fallback} mechanism, which allows the system to continue answering with subsequent retrieved candidates when the current table lacks sufficient evidence. We compare the full model with two variants, \texttt{w/o fallback} and \texttt{w/o refine}, under both 5-step and 10-step budgets using Qwen3-30B on TCR-Bench (Mixed format). In addition, we evaluate two variants to study the effect of our \texttt{Structure-Preserving} design: \texttt{match-only}, where refinement retains only matched rows without preserving table structure, and \texttt{match-only QA}, where both refinement and final QA operate only on matched rows. More fine-grained ablations are provided in the appendix.

\begin{table}[t]
\centering
\small
\begin{tabular}{lcccc}
\toprule
Strategy & R@all & F1 & Avg.\ Steps & Time (h) \\
\midrule
\textbf{5-step setting} & & & & \\
\midrule
5-step (Full) & \textbf{0.416} & \textbf{0.284} & 4.407 & 2.335 \\
w/o fallback & 0.364 & 0.263 & 3.919 & 2.094 \\
w/o refine & 0.239 & 0.173 & 3.555 & 1.733 \\
match-only & 0.359 & 0.249 & 4.435 & 1.841 \\
match-only QA & 0.359 & 0.065 & 4.445 & 1.132 \\
\midrule
\textbf{10-step setting} & & & & \\
\midrule
10-step (Full) & \textbf{0.507} & \textbf{0.361} & 6.244 & 3.466 \\
w/o fallback & 0.426 & 0.311 & 4.689 & 2.616 \\
w/o refine & 0.268 & 0.195 & 5.019 & 2.408 \\
match-only & 0.440 & 0.316 & 6.421 & 2.941 \\
match-only QA & 0.445 & 0.080 & 6.096 & 2.561 \\
\bottomrule
\end{tabular}
\caption{Ablation study on refine, fallback, and Structure-Preserving mechanisms on TCR-Bench (Mixed format) using Qwen3-30B.}
\label{tab:component_ablation}
\end{table}

Table~\ref{tab:component_ablation} shows that both refine and fallback contribute positively, with refine providing the dominant improvement. Removing refine substantially reduces performance, causing R@all/F1 to drop from 0.416/0.284 to 0.239/0.173 under 5 steps, and from 0.507/0.361 to 0.268/0.195 under 10 steps. Removing fallback also leads to a consistent decline, indicating that allowing the system to continue QA on subsequently retrieved candidates improves robustness when the current table lacks sufficient evidence.

The match-only variants further demonstrate the value of the \texttt{Structure-Preserving} design. Retaining only matched rows consistently underperforms the full model, reducing R@all from 0.416 to 0.359 under 5 steps and from 0.507 to 0.440 under 10 steps. When table structure is removed from both refinement and QA (match-only QA), F1 drops sharply to 0.065 and 0.080 despite similar retrieval performance. These results suggest that the \texttt{Structure-Preserving} design provides useful schema-level context for both retrieval and downstream answer generation.

\subsection{Effect of Iterative Refinement on Retrieval Quality}
\label{subsec:refine-rounds}

To analyze the effect of iterative refinement itself, we track whether the top-1 retrieved table exactly matches the gold table across successive \texttt{refine-regenerate} rounds, where round 0 corresponds to the initial \texttt{pattern-gen} output. Early-terminated samples retain their last observed hit status in later rounds, allowing each round to be evaluated over the full test set. This analysis focuses on the quality of the generated search patterns before the final table selection step.

\begin{figure}[t]
    \centering
    \includegraphics[width=0.95\linewidth]{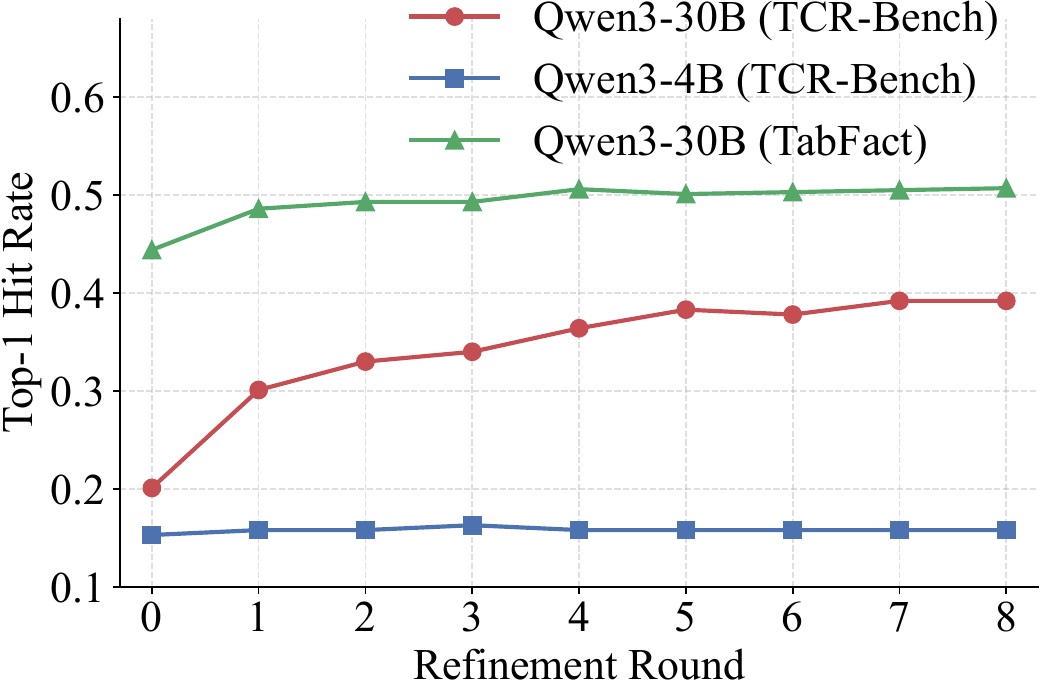}
    \caption{Top-1 hit rate as a function of refinement round. Iterative refinement continues to improve retrieval quality for the stronger Qwen3-30B model well beyond the first round, whereas the smaller Qwen3-4B model plateaus almost immediately, indicating that the marginal benefit of refinement scales with model capability.}
    \label{fig:refine-rounds}
\end{figure}

Figure~\ref{fig:refine-rounds} shows that iterative refinement improves evidence discovery beyond the initial pattern generation, especially for the stronger Qwen3-30B model. While both models obtain most gains in the first few rounds, Qwen3-30B continues improving until later rounds, whereas Qwen3-4B quickly saturates. This indicates that the benefit of additional refinement depends on the reasoning capability of the underlying model. Full per-round results are provided in the appendix.

\subsubsection{Action Usage Breakdown}
\label{subsec:action_breakdown}
To understand how \ours{} utilizes its search budget, we record the average number of times each action type is executed per query. \texttt{refine-select} and \texttt{refine-regenerate} denote selecting among candidate tables and regenerating search patterns, respectively, while QA (total) counts all QA calls.

\begin{figure}[t]
\centering
\includegraphics[width=0.95\linewidth]{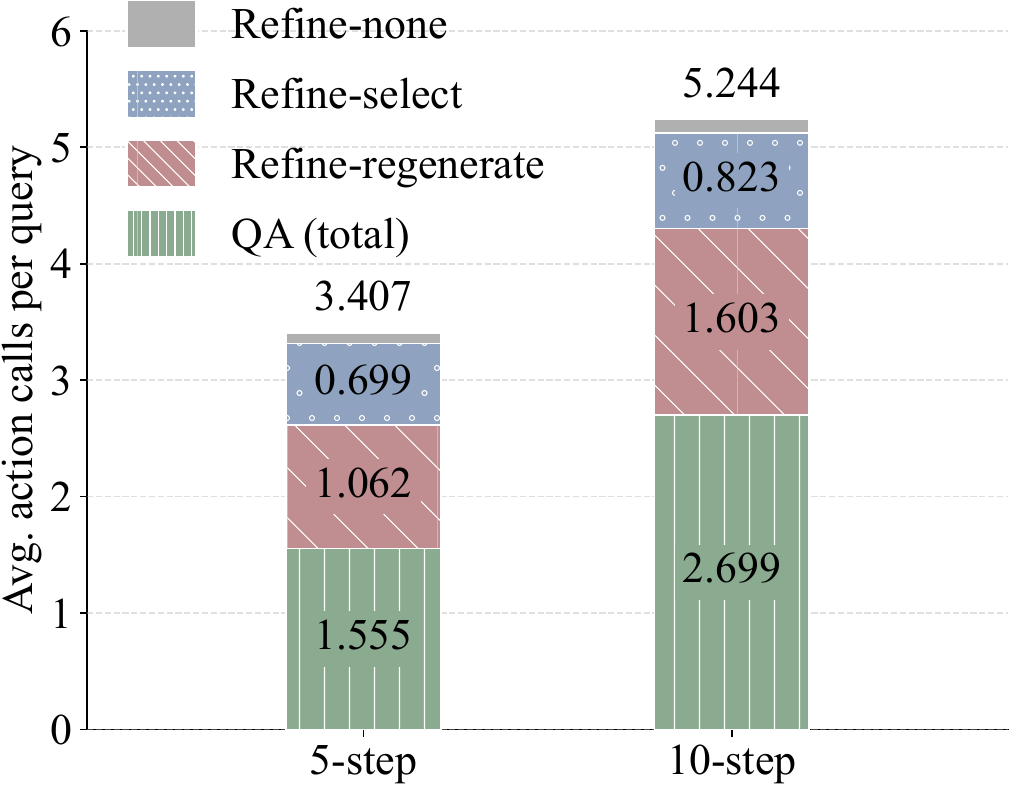}
\caption{Average action counts per query on TCR-Bench for the default \ours{} configuration, excluding the initial `\texttt{pattern-gen}` step (fixed to 1 per query).
}
\label{fig:action_breakdown}
\end{figure}

Figure~\ref{fig:action_breakdown} shows that \texttt{refine-regenerate} is consistently invoked more frequently than \texttt{refine-select} (1.062 vs. 0.699 at 5 steps and 1.603 vs. 0.823 at 10 steps). This suggests that, when additional refinement is performed, the agent more often improves retrieval by revising search patterns than by relying on candidate-table selection. Increasing the step budget primarily leads to more refinement and QA actions, indicating that the extra computation is allocated to continued evidence exploration and validation. These action statistics provide a \textbf{transparent} view of the agent's behavior, showing that additional budget is predominantly used to generate and evaluate new retrieval hypotheses through pattern revision.

\subsection{Case Study: Refine Behavior}
\label{subsec:refine-case-study}

To provide a more \textbf{transparent} view of how \ours{} improves retrieval, we conduct a qualitative analysis of the refine process. Because the agent explicitly generates search patterns and exposes its intermediate reasoning steps, its retrieval decisions can be inspected and analyzed directly, facilitating both diagnosis and future system improvements. Since QA generation follows the standard benchmark protocol, we focus on retrieval refinement behaviors; detailed examples are provided in the appendix.

Successful refinement mainly occurs in two situations. First, when query intent is clear but schema terminology differs from corpus conventions, the agent adapts its search terms to schema names observed in retrieved candidates. Second, queries containing distinctive evidence values often trigger a transition from schema-oriented matching to value-aware or hybrid retrieval. These behaviors help \ours{} identify evidence-bearing tables despite schema variation and heterogeneous naming conventions.

Failures primarily occur when multiple highly related candidate tables coexist in the corpus. In such cases, despite the evidence-oriented refinement strategy, the underlying model can still be influenced by strong semantic relevance signals and prematurely converge on a plausible candidate before sufficient evidence verification is completed. Queries requiring deeper schema exploration or alternative value representations may also expose limited refinement persistence. These observations align with our quantitative analyses, suggesting that iterative evidence seeking improves retrieval quality while remaining constrained by the reasoning capability of the underlying model.

These observations suggest that future improvements may come from stronger refinement policies and more reliable evidence-verification mechanisms rather than from additional semantic matching capacity.

\section{Conclusion}

We introduced \ours{}, a structure-preserving agentic search framework for open-domain table retrieval. By iteratively refining search patterns and inspecting evidence rather than relying solely on static similarity matching, \ours{} shifts retrieval toward evidence sufficiency and answerability. Experiments on heterogeneous table benchmarks demonstrate competitive performance against strong retrieval-and-reranking baselines, while requiring neither retriever training nor corpus-level semantic indexing.

Our analyses show iterative refinement drives \ours{}'s performance gains by adapting to heterogeneous schemas, but its effectiveness depends on LLM reasoning and comes at multi-interaction cost.

Overall, our results suggest that active evidence seeking can complement traditional similarity-based retrieval for tabular tasks. We hope this work encourages further exploration of agentic retrieval strategies that combine structured evidence exploration with conventional retrieval techniques.


\bibliography{aaai2027}

@article{DCI,
  title={Beyond semantic similarity: Rethinking retrieval for agentic search via direct corpus interaction},
  author={Li, Zhuofeng and Zhang, Haoxiang and Wei, Cong and Lu, Pan and Nie, Ping and Lu, Yi and Bai, Yuyang and Feng, Shangbin and Zhu, Hangxiao and Zhong, Ming and others},
  journal={arXiv preprint arXiv:2605.05242},
  year={2026}
}

@article{GrepRAG,
  title={GrepRAG: An Empirical Study and Optimization of Grep-Like Retrieval for Code Completion},
  author={Wang, Baoyi and Wang, Xingliang and Li, Guochang and Zhi, Chen and Han, Junxiao and Zhao, Xinkui and Wang, Nan and Deng, Shuiguang and Yin, Jianwei},
  journal={arXiv preprint arXiv:2601.23254},
  year={2026}
}

@inproceedings{NQ-TABLES-DTR,
  title={Open domain question answering over tables via dense retrieval},
  author={Herzig, Jonathan and M{\"u}ller, Thomas and Krichene, Syrine and Eisenschlos, Julian},
  booktitle={Proceedings of the 2021 Conference of the North American Chapter of the Association for Computational Linguistics: Human Language Technologies},
  pages={512--519},
  year={2021}
}

@inproceedings{DPR,
  title={Dense passage retrieval for open-domain question answering},
  author={Karpukhin, Vladimir and Oguz, Barlas and Min, Sewon and Lewis, Patrick and Wu, Ledell and Edunov, Sergey and Chen, Danqi and Yih, Wen-tau},
  booktitle={Proceedings of the 2020 conference on empirical methods in natural language processing (EMNLP)},
  pages={6769--6781},
  year={2020}
}

@inproceedings{table-specific-embedding-not-better,
  title={Table retrieval may not necessitate table-specific model design},
  author={Wang, Zhiruo and Jiang, Zhengbao and Nyberg, Eric and Neubig, Graham},
  booktitle={Proceedings of the workshop on structured and unstructured knowledge integration (SUKI)},
  pages={36--46},
  year={2022}
}

@inproceedings{THoRR,
  title={THoRR: Complex Table Retrieval and Refinement for RAG.},
  author={Kim, Kihun and Kim, Mintae and Lee, Hokyung and Park, Seong Ik and Han, Youngsub and Jeon, Byoung-Ki},
  booktitle={IR-RAG@ SIGIR},
  pages={50--55},
  year={2024}
}

@inproceedings{TEM,
  title={Tabular embedding model (tem): Finetuning embedding models for tabular rag applications},
  author={Khanna, Sujit and Subedi, Shishir},
  booktitle={Intelligent Computing-Proceedings of the Computing Conference},
  pages={448--460},
  year={2025},
  organization={Springer}
}

@inproceedings{THYME,
  title={Tailoring Table Retrieval from a Field-aware Hybrid Matching Perspective},
  author={Li, Da and Bi, Keping and Guo, Jiafeng and Cheng, Xueqi},
  booktitle={Proceedings of the 2025 Conference on Empirical Methods in Natural Language Processing},
  pages={27681--27692},
  year={2025}
}

@inproceedings{CRAFT,
  title={CRAFT: Training-Free Cascaded Retrieval for Tabular QA},
  author={Singh, Adarsh and Bhandari, Kushal Raj and Gao, Jianxi and Dan, Soham and Gupta, Vivek},
  booktitle={Proceedings of the 64th Annual Meeting of the Association for Computational Linguistics (Volume 1: Long Papers)},
  pages={3284--3298},
  year={2026}
}

@inproceedings{JAR,
  title={Is table retrieval a solved problem? exploring join-aware multi-table retrieval},
  author={Chen, Peter Baile and Zhang, Yi and Roth, Dan},
  booktitle={Proceedings of the 62nd annual meeting of the association for computational linguistics (volume 1: long papers)},
  pages={2687--2699},
  year={2024}
}

@inproceedings{TARGET,
  title={TARGET: Benchmarking Table Retrieval for Generative Tasks},
  author={Ji, Xingyu and Parameswaran, Aditya and Hulsebos, Madelon},
  booktitle={NeurIPS 2024 Third Table Representation Learning Workshop},
  year={2024}
}

@inproceedings{t2-ragbench,
  title={T2-RAGBench: Text-and-Table Aware Retrieval-Augmented Generation},
  author={Strich, Jan and Isgorur, Enes Kutay and Trescher, Maximilian and Biemann, Chris and Semmann, Martin},
  booktitle={Proceedings of the 19th Conference of the European Chapter of the Association for Computational Linguistics (Volume 1: Long Papers)},
  pages={165--191},
  year={2026}
}

@article{MultiTableQA,
  title={RAG over Tables: Hierarchical Memory Index, Multi-Stage Retrieval, and Benchmarking},
  author={Zou, Jiaru and Fu, Dongqi and Chen, Sirui and He, Xinrui and Li, Zihao and Zhu, Yada and Han, Jiawei and He, Jingrui},
  journal={arXiv preprint arXiv:2504.01346},
  year={2025}
}

@article{qwen3embedding,
  title={Qwen3 Embedding: Advancing Text Embedding and Reranking Through Foundation Models},
  author={Zhang, Yanzhao and Li, Mingxin and Long, Dingkun and Zhang, Xin and Lin, Huan and Yang, Baosong and Xie, Pengjun and Yang, An and Liu, Dayiheng and Lin, Junyang and Huang, Fei and Zhou, Jingren},
  journal={arXiv preprint arXiv:2506.05176},
  year={2025}
}

@misc{Stella,
      title={Jasper and Stella: distillation of SOTA embedding models}, 
      author={Dun Zhang and Jiacheng Li and Ziyang Zeng and Fulong Wang},
      year={2025},
      eprint={2412.19048},
      archivePrefix={arXiv},
      primaryClass={cs.IR},
      url={https://arxiv.org/abs/2412.19048}, 
}

@misc{jina-embeddings-v4,
      title={jina-embeddings-v4: Universal Embeddings for Multimodal Multilingual Retrieval}, 
      author={Michael Günther and Saba Sturua and Mohammad Kalim Akram and Isabelle Mohr and Andrei Ungureanu and Sedigheh Eslami and Scott Martens and Bo Wang and Nan Wang and Han Xiao},
      year={2025},
      eprint={2506.18902},
      archivePrefix={arXiv},
      primaryClass={cs.AI},
      url={https://arxiv.org/abs/2506.18902}, 
}

@misc{qwen3technicalreport,
      title={Qwen3 Technical Report}, 
      author={Qwen Team},
      year={2025},
      eprint={2505.09388},
      archivePrefix={arXiv},
      primaryClass={cs.CL},
      url={https://arxiv.org/abs/2505.09388}, 
}

@article{LLaMA,
  title={The llama 3 herd of models},
  author={Grattafiori, Aaron and Dubey, Abhimanyu and Jauhri, Abhinav and Pandey, Abhinav and Kadian, Abhishek and Al-Dahle, Ahmad and Letman, Aiesha and Mathur, Akhil and Schelten, Alan and Vaughan, Alex and others},
  journal={arXiv preprint arXiv:2407.21783},
  year={2024}
}

@article{Agentic_RAG-survey,
  title={Agentic retrieval-augmented generation: A survey on agentic rag},
  author={Singh, Aditi and Ehtesham, Abul and Kumar, Saket and Khoei, Tala Talaei and Vasilakos, Athanasios V},
  journal={arXiv preprint arXiv:2501.09136},
  year={2025}
}

@inproceedings{REACT,
  title={REACT: SYNERGIZING REASONING AND ACTING IN LANGUAGE MODELS},
  author={Yao, Shunyu and Zhao, Jeffrey and Yu, Dian and Du, Nan and Shafran, Izhak and Narasimhan, Karthik and Cao, Yuan},
  booktitle={11th International Conference on Learning Representations, ICLR 2023},
  year={2023}
}

@inproceedings{Search-o1,
  title={Search-o1: Agentic search-enhanced large reasoning models},
  author={Li, Xiaoxi and Dong, Guanting and Jin, Jiajie and Zhang, Yuyao and Zhou, Yujia and Zhu, Yutao and Zhang, Peitian and Dou, Zhicheng},
  booktitle={Proceedings of the 2025 Conference on Empirical Methods in Natural Language Processing},
  pages={5420--5438},
  year={2025}
}

@inproceedings{RAG-more-context2-Re-rag,
  title={Re-rag: Improving open-domain qa performance and interpretability with relevance estimator in retrieval-augmented generation},
  author={Kim, Kiseung and Lee, Jay-Yoon},
  booktitle={Proceedings of the 2024 Conference on Empirical Methods in Natural Language Processing},
  pages={22149--22161},
  year={2024}
}

@inproceedings{RAG-more-context1,
  title={Making Retrieval-Augmented Language Models Robust to Irrelevant Context},
  author={Yoran, Ori and Wolfson, Tomer and Ram, Ori and Berant, Jonathan},
  booktitle={ICLR 2024 Workshop on Large Language Model (LLM) Agents},
  year = {2024}
}

@article{RAG-more-context3-lost-in-middle,
  title={Dynamic Context Selection for Retrieval-Augmented Generation: Mitigating Distractors and Positional Bias},
  author={Iratni, Malika and Boughanem, Mohand and Dkaki, Taoufiq},
  journal={arXiv preprint arXiv:2512.14313},
  year={2025}
}

@inproceedings{RAG-more-context4,
  title={Hoh: A dynamic benchmark for evaluating the impact of outdated information on retrieval-augmented generation},
  author={Ouyang, Jie and Pan, Tingyue and Cheng, Mingyue and Yan, Ruiran and Luo, Yucong and Lin, Jiaying and Liu, Qi},
  booktitle={Proceedings of the 63rd Annual Meeting of the Association for Computational Linguistics (Volume 1: Long Papers)},
  pages={6036--6063},
  year={2025}
}

@article{FeTaQA,
  title={FeTaQA: Free-form table question answering},
  author={Nan, Linyong and Hsieh, Chiachun and Mao, Ziming and Lin, Xi Victoria and Verma, Neha and Zhang, Rui and Kry{\'s}ci{\'n}ski, Wojciech and Schoelkopf, Hailey and Kong, Riley and Tang, Xiangru and others},
  journal={Transactions of the Association for Computational Linguistics},
  volume={10},
  pages={35--49},
  year={2022},
  publisher={MIT Press One Broadway, 12th Floor, Cambridge, Massachusetts 02142, USA~…}
}

@inproceedings{OTTQA,
  title={Open Question Answering over Tables and Text},
  author={Chen, Wenhu and Chang, Ming-Wei and Schlinger, Eva and Wang, William Yang and Cohen, William W},
  year={2021},
  booktitle={International Conference on Learning Representations}
}

@book{BM25,
  title     = {The Probabilistic Relevance Framework: BM25 and Beyond},
  author    = {Stephen E. Robertson and Hugo Zaragoza},
  publisher = {Now Publishers Inc.},
  year      = {2009},
  isbn      = {9781601981919},
  note      = {Foundations and Trends in Information Retrieval},
}

@article{TCR-Bench,
  title={Semantically Similar, Logically Distinct: Diagnosing the Semantic-Answerability Gap in Table RAG},
  author={Tian, Jiaming and Li, Liyao and Ye, Wentao and Wang, Haobo and Yu, Lihua and Ren, Zujie and Chen, Gang and Zhao, Junbo},
  journal={arXiv preprint arXiv:2607.17742},
  year={2026}
}

@inproceedings{WikiTQ,
  title={Compositional semantic parsing on semi-structured tables},
  author={Pasupat, Panupong and Liang, Percy},
  booktitle={Proceedings of the 53rd Annual Meeting of the Association for Computational Linguistics and the 7th International Joint Conference on Natural Language Processing (Volume 1: Long Papers)},
  pages={1470--1480},
  year={2015}
}

@inproceedings{HybridQA,
  title={HybridQA: A dataset of multi-hop question answering over tabular and textual data},
  author={Chen, Wenhu and Zha, Hanwen and Chen, Zhiyu and Xiong, Wenhan and Wang, Hong and Wang, William Yang},
  booktitle={Findings of the Association for Computational Linguistics: EMNLP 2020},
  pages={1026--1036},
  year={2020}
}

@inproceedings{TAPEX,
  title={TAPEX: Table Pre-training via Learning a Neural SQL Executor},
  author={Liu, Qian and Chen, Bei and Guo, Jiaqi and Ziyadi, Morteza and Lin, Zeqi and Chen, Weizhu and Lou, Jian-Guang},
  booktitle={International Conference on Learning Representations},
  year={2022}
}

@inproceedings{Structgpt,
  title={Structgpt: A general framework for large language model to reason over structured data},
  author={Jiang, Jinhao and Zhou, Kun and Dong, Zican and Ye, Keming and Zhao, Xin and Wen, Ji-Rong},
  booktitle={Proceedings of the 2023 conference on empirical methods in natural language processing},
  pages={9237--9251},
  year={2023}
}

@inproceedings{Colbert,
  title={Colbert: Efficient and effective passage search via contextualized late interaction over bert},
  author={Khattab, Omar and Zaharia, Matei},
  booktitle={Proceedings of the 43rd International ACM SIGIR conference on research and development in Information Retrieval},
  pages={39--48},
  year={2020}
}

@article{Contriever,
  title={Unsupervised Dense Information Retrieval with Contrastive Learning},
  author={Izacard, Gautier and Caron, Mathilde and Hosseini, Lucas and Riedel, Sebastian and Bojanowski, Piotr and Joulin, Armand and Grave, Edouard},
  journal={Transactions on Machine Learning Research},
  year={2022}
}

@inproceedings{TabFact,
  title={TabFact: A Large-scale Dataset for Table-based Fact Verification},
  author={Chen, Wenhu and Wang, Hongmin and Chen, Jianshu and Zhang, Yunkai and Wang, Hong and Li, Shiyang and Zhou, Xiyou and Wang, William Yang},
  booktitle={International Conference on Learning Representations},
  year={2020}
}

@article{TRACK-RANK-CRACK,
  title={Track, Rank, Crack: Epistemic Working Memory Scales Multi-Hop Reasoning in Language Agents},
  author={Liu, Ning},
  journal={arXiv preprint arXiv:2607.12267},
  year={2026}
}

@inproceedings{ACE,
  title={Resolving evidence sparsity: Agentic context engineering for long-document understanding},
  author={Liu, Keliang and Chen, Zizhi and Li, Mingcheng and Tang, Jingqun and Yang, Dingkang and Zhang, Lihua},
  booktitle={Proceedings of the IEEE/CVF Conference on Computer Vision and Pattern Recognition},
  pages={19452--19462},
  year={2026}
}

@inproceedings{Self-RAG,
  title={Self-rag: Learning to retrieve, generate, and critique through self-reflection},
  author={Asai, Akari and Wu, Zeqiu and Wang, Yizhong and Sil, Avi and Hajishirzi, Hannaneh},
  booktitle={International conference on learning representations},
  volume={2024},
  pages={9112--9141},
  year={2024}
}

@article{BIRD,
  title={Can llm already serve as a database interface? a big bench for large-scale database grounded text-to-sqls},
  author={Li, Jinyang and Hui, Binyuan and Qu, Ge and Yang, Jiaxi and Li, Binhua and Li, Bowen and Wang, Bailin and Qin, Bowen and Geng, Ruiying and Huo, Nan and others},
  journal={Advances in Neural Information Processing Systems},
  volume={36},
  year={2024}
}

@inproceedings{Spider,
  title={Spider: A large-scale human-labeled dataset for complex and cross-domain semantic parsing and text-to-SQL task},
  author={Yu, Tao and Zhang, Rui and Yang, Kai and Yasunaga, Michihiro and Wang, Dongxu and Li, Zifan and Ma, James and Li, Irene and Yao, Qingning and Roman, Shanelle and others},
  booktitle={Proceedings of the 2018 Conference on Empirical Methods in Natural Language Processing},
  year={2018}
}

@inproceedings{SPLADE,
  title={Splade: Sparse lexical and expansion model for first stage ranking},
  author={Formal, Thibault and Piwowarski, Benjamin and Clinchant, St{\'e}phane},
  booktitle={Proceedings of the 44th International ACM SIGIR Conference on Research and Development in Information Retrieval},
  pages={2288--2292},
  year={2021}
}

@article{SPLADE-v2,
  title={SPLADE v2: Sparse lexical and expansion model for information retrieval},
  author={Formal, Thibault and Lassance, Carlos and Piwowarski, Benjamin and Clinchant, St{\'e}phane},
  journal={arXiv preprint arXiv:2109.10086},
  year={2021}
}

@article{SPLADE-v3,
  title={SPLADE-v3: New baselines for SPLADE},
  author={Lassance, Carlos and D{\'e}jean, Herv{\'e} and Formal, Thibault and Clinchant, St{\'e}phane},
  journal={arXiv preprint arXiv:2403.06789},
  year={2024}
}

@article{Jina-reranker-v3,
  title={Jina-reranker-v3: Last but Not Late Interaction for Listwise Document Reranking},
  author={Wang, Feng and Li, Yuqing and Xiao, Han},
  journal={arXiv preprint arXiv:2509.25085},
  year={2025}
}

@misc{SauerkrautLM-Multi-ModernColBERT,
  title={SauerkrautLM-Multi-ModernColBERT},
  author={David Golchinfar},
  url={https://huggingface.co/VAGOsolutions/SauerkrautLM-Multi-ModernColBERT},
  year={2025}
}

@misc{GLM-Z1,
      title={ChatGLM: A Family of Large Language Models from GLM-130B to GLM-4 All Tools}, 
      author={Team GLM and Aohan Zeng and Bin Xu and Bowen Wang and Chenhui Zhang and Da Yin and Diego Rojas and Guanyu Feng and Hanlin Zhao and Hanyu Lai and Hao Yu and Hongning Wang and Jiadai Sun and Jiajie Zhang and Jiale Cheng and Jiayi Gui and Jie Tang and Jing Zhang and Juanzi Li and Lei Zhao and Lindong Wu and Lucen Zhong and Mingdao Liu and Minlie Huang and Peng Zhang and Qinkai Zheng and Rui Lu and Shuaiqi Duan and Shudan Zhang and Shulin Cao and Shuxun Yang and Weng Lam Tam and Wenyi Zhao and Xiao Liu and Xiao Xia and Xiaohan Zhang and Xiaotao Gu and Xin Lv and Xinghan Liu and Xinyi Liu and Xinyue Yang and Xixuan Song and Xunkai Zhang and Yifan An and Yifan Xu and Yilin Niu and Yuantao Yang and Yueyan Li and Yushi Bai and Yuxiao Dong and Zehan Qi and Zhaoyu Wang and Zhen Yang and Zhengxiao Du and Zhenyu Hou and Zihan Wang},
      year={2024},
      eprint={2406.12793},
      archivePrefix={arXiv},
      primaryClass={id='cs.CL' full_name='Computation and Language' is_active=True alt_name='cmp-lg' in_archive='cs' is_general=False description='Covers natural language processing. Roughly includes material in ACM Subject Class I.2.7. Note that work on artificial languages (programming languages, logics, formal systems) that does not explicitly address natural-language issues broadly construed (natural-language processing, computational linguistics, speech, text retrieval, etc.) is not appropriate for this area.'}
}

@article{TableGPT-R1,
  title={TableGPT-R1: Advancing Tabular Reasoning Through Reinforcement Learning},
  author={Yang, Saisai and Huang, Qingyi and Yuan, Jing and Zha, Liangyu and Tang, Kai and Yang, Yuhang and Wang, Ning and Wei, Yucheng and Li, Liyao and Ye, Wentao and others},
  journal={arXiv preprint arXiv:2512.20312},
  year={2025}
}


\appendix

\FloatBarrier


\section{Overview of the Appendix}
This appendix provides additional details and analyses that support the main paper. The contents are organized as follows:

\begin{itemize}
    \item \textbf{Algorithm Details} (Section \S \textit{``Algorithm of \ours{}''}): A complete description of our proposed algorithm.
    \item \textbf{Experimental Setup} (Section \S \textit{``Additional Experimental Details''}): Detailed configurations and implementation specifics of our experiments.
    \item \textbf{Hyperparameters} (Section \S \textit{``Hyperparameters''}): Detailed configuration of key hyperparameters.
    \item \textbf{Prompts Used} (Section \S \textit{``Prompts Used in Our Pipeline''}): All prompt templates employed in our framework.
    \item \textbf{Extended Main Results} (Section \S \textit{``Additional Backbone Analysis''}): Additional experimental results and analyses expanding on the main experiments.
    \item \textbf{Oracle Performance} (Section \S \textit{``Oracle Results (Gold Table Only)''}): Results using the gold table as input for TableQA, serving as a reference upper bound.
     \item \textbf{Additional Retrieval-Only Results} (Section \S \textit{``Additional Retrieval-Only Results''}): A summary of retrieval-only method performances without the QA stage.
    \item \textbf{Discriminative Score Analysis} (Section \S \textit{``Details of Discriminative Score Metrics''}): In-depth discussion on the Discriminative Score introduced in the main paper, along with detailed specifications of the metrics employed.
    \item \textbf{Efficiency Analysis} (Section \S \textit{``Full Efficiency Breakdown: Calls and Runtime''}): Detailed efficiency and computational cost analysis.
    \item \textbf{Fine-Grained Ablation} (Section \S \textit{``Fine-Grained Ablation of the Refine Mechanism''}): Fine-grained ablation of the refine mechanism.
    \item \textbf{Refinement Iteration Analysis} (Section \S \textit{``Sustained Improvement Across Rounds''}): Performance trends of the retrieval process across refinement rounds.
    \item \textbf{Refinement Strategy Analysis} (Section \S \textit{``Refine Strategies''}): Fine-grained examination of different refinement strategies.
    \item \textbf{Robustness Analysis} (Section \S \textit{``Detailed Results for Table Format Robustness''}): Comprehensive robustness evaluation under various conditions, with detailed quantitative results, multiple independent runs for statistical significance, and a range of refinement strategies to assess performance stability across different configurations.
    \item \textbf{Effect of Retrieved Evidence Size} (Section \S \textit{``Effect of Retrieved Evidence Size''}): Examines the trade-off between evidence completeness and efficiency by varying the amount of table content exposed to the LLM during retrieval, with detailed performance comparisons and cost analysis across different evidence-size configurations.
    \item \textbf{Transposed Table Stress Test} (Section \S \textit{``Transposed Tables: A Stress Test Beyond Row-Oriented Entity Structure''}): Stress-test results on transposed TCR-Bench, where column-distributed attributes disrupt row-oriented heuristics.
    \item \textbf{Full Case Study} (Section \S \textit{``Complete Case Study: Refine Behavior''}): Complete case examples illustrating our method's behavior.
    \item \textbf{Limitations} (Section \S \textit{``Limitations''}): Discussion of the key limitations of our work, including model generalizability, lack of optimization techniques, and benchmark scale constraints.
    \item \textbf{Ethical Statement} (Section \S \textit{``Ethical Statement''}): Statement on data usage, potential risks of output bias, and the role of generative AI tools in our research.
\end{itemize}

\section{Algorithm of \ours{}}
\label{sec:supp-algorithm}

We provide the complete inference procedure of \ours{} in Algorithm~\ref{alg:ours}. The algorithm describes the iterative evidence-seeking process, including initial retrieval, agentic refinement, fallback selection, and answer verification.

\begin{algorithm}[tb]
\caption{Iterative Evidence-Seeking Procedure of \ours{}}
\label{alg:ours}
\begin{algorithmic}[1]
\REQUIRE Query $q$; table store $\mathcal{S}$; task type $\tau$; hyperparameters $n$, $K_{\mathrm{fb}}$, $S_{\max}$
\ENSURE Retrieved tables $\mathcal{K}$, answer $\hat{y}$

\STATE $\mathcal{P} \leftarrow \mathrm{LLM\_GeneratePatterns}(q,\tau)$; $s \leftarrow 1$
\STATE $\mathcal{R} \leftarrow \mathrm{RankTables}(\mathcal{S},\mathcal{P})$

\WHILE{$s < S_{\max}-1$}
    \STATE Build previews for top-$n$ retrieved tables in $\mathcal{R}$
    \STATE $\delta \leftarrow \mathrm{LLM\_RefineDecision}(q,\tau,\text{previews},\mathcal{P})$
    \STATE $s \leftarrow s+1$
    
    \IF{$\delta.\mathit{action}=\texttt{select}$}
        \STATE $\mathcal{K}\leftarrow\delta.\mathit{selected\_tables}$; \textbf{break}
    \ELSIF{$\delta.\mathit{action}=\texttt{regenerate}$}
        \STATE $\mathcal{P}\leftarrow\delta.\mathit{patterns}$
        \STATE $\mathcal{R}\leftarrow\mathrm{RankTables}(\mathcal{S},\mathcal{P})$
    \ELSE
        \STATE \textbf{break}
    \ENDIF
\ENDWHILE

\IF{$\mathcal{K}$ is empty}
    \STATE $\mathcal{K}\leftarrow\{T_{(1)},\ldots,T_{(K_{\mathrm{fb}})}\}$ from $\mathcal{R}$
\ENDIF

\STATE $\hat{y}\leftarrow\mathrm{LLM\_QA}(q,C_{\mathcal{K}},\tau)$; $s\leftarrow s+1$

\FOR{each $T\in\mathcal{R}\setminus\mathcal{K}$ while $\hat{y}$ is unsatisfactory and $s<S_{\max}$}
    \STATE $\hat{y}\leftarrow\mathrm{LLM\_QA}(q,C_{\{T\}},\tau)$
    \STATE $s\leftarrow s+1$
\ENDFOR

\RETURN $\mathcal{K}$, $\hat{y}$
\end{algorithmic}
\end{algorithm}

\subsection{Why Pattern-Based Retrieval Instead of Direct Table Navigation}
\label{app:retrieval_interface}

A natural question is why \ours{} performs retrieval through generated search patterns rather than allowing the agent to directly browse or select tables from the repository.

This design largely follows the retrieval protocol adopted by existing open-domain table retrieval benchmarks. In TCR-Bench, for example, retrieval methods operate over table content while the underlying storage metadata is not exposed to the model. This separation is important because filenames, table titles, or repository-specific identifiers may contain information that partially reveals the answer-bearing table, introducing unintended shortcuts that bypass evidence-based retrieval.

Accordingly, our implementation only exposes retrieval results and temporary anonymous identifiers used for interaction during the retrieval process. These identifiers serve purely as references for agent actions and do not correspond to corpus metadata visible to the model. As a result, the agent cannot navigate the repository through filenames, storage paths, or other dataset-specific metadata, and must instead interact with the corpus through content-based retrieval operations.

This design is also consistent with the nature of heterogeneous table repositories. In practice, tables may originate from multiple sources and formats, while benchmark datasets often adopt a unified storage format primarily for implementation convenience. For example, the CSV files provided in TCR-Bench should be viewed as an underlying storage representation rather than the natural form of interaction available to a retrieval system. To maintain a comparable retrieval setting, we adopt the same abstraction when applying \ours{} to TARGET and expose only retrieval-accessible content rather than storage-level metadata.

Within this setting, search patterns provide a simple and transparent retrieval interface. They allow the agent to formulate retrieval hypotheses, inspect returned evidence, and iteratively adjust subsequent searches based on observed schema conventions, value representations, and retrieval failures. The resulting process therefore resembles evidence-guided corpus interaction rather than direct repository navigation.

We emphasize that this choice is motivated by the evaluation setting rather than by any assumption that pattern-based retrieval is the only possible interface. Alternative retrieval interfaces could potentially be incorporated into the same evidence-seeking framework, provided that they support iterative interaction and evidence-driven refinement.

\section{Additional Experimental Details}
\label{sec:Additional-Experimental-Details}

\subsection{Embedding and Reranker Models}

\textbf{Qwen3-Embedding-4B (E4B).}
Qwen3-Embedding-4B~\citep{qwen3embedding} is an instruction-aware embedding model supporting contexts up to 32k tokens, multilingual retrieval, and Matryoshka Representation Learning for flexible embedding dimensions. We use the 4B variant throughout the main experiments.

\textbf{Stella.}
\texttt{stella\_en\_1.5B\_v5}~\citep{Stella} is a 1.5B-parameter English embedding model optimized for semantic similarity and dense retrieval. It serves as a competitive mid-scale dense retriever baseline.

\textbf{Jina.}
\texttt{jina-embeddings-v4}~\citep{jina-embeddings-v4} is a multilingual and multimodal embedding model with approximately 3.8B parameters. It supports both single-vector dense retrieval and multi-vector late-interaction retrieval. We include it in additional experiments reported later in this appendix.

\textbf{Qwen3-Reranker-0.6B (R0.6B).}
Qwen3-Reranker-0.6B~\citep{qwen3embedding} is a lightweight reranking model with 0.6B parameters. Despite its compact size, it delivers effective relevance scoring and serves as an efficient option for latency-sensitive reranking scenarios.

\textbf{Qwen3-Reranker-4B (R4B).}
Qwen3-Reranker-4B~\citep{qwen3embedding} is adopted as our primary reranking model. We evaluate its performance by reranking the top-3 and top-5 candidates retrieved by the dense retriever.

\textbf{Jina-Reranker-v3.}
\texttt{jina-reranker-v3}~\citep{Jina-reranker-v3} is a 0.6B multilingual reranker with a novel "last but not late" interaction architecture. Unlike Qwen3-Reranker, it processes query and documents in the same context window via causal self-attention, supporting up to 64 documents at once within a 131K token context.

\subsection{Alternative Retrieval Paradigms}

To complement the dense approaches discussed above, we also evaluate three representative baselines from alternative paradigms.

\textbf{BM25~\citep{BM25}.}
A classic sparse retrieval baseline based on term frequency, inverse document frequency, and length normalization. It remains competitive due to its efficiency and interpretability, with no learnable parameters.

\textbf{Splade~\citep{SPLADE, SPLADE-v2} (Splade-v3~\citep{SPLADE-v3}).}
A learned sparse retrieval model that uses a transformer encoder to generate sparse term weight vectors, simultaneously performing document expansion and term weighting. It preserves the inverted-index compatibility of sparse retrieval while mitigating the vocabulary mismatch problem of BM25.

\textbf{ColBERT~\citep{Colbert} (SauerkrautLM-Multi-ModernColBERT~\citep{SauerkrautLM-Multi-ModernColBERT}).}
A late-interaction retrieval model that encodes queries and documents independently and computes relevance via token-level MaxSim operations.

\subsection{Backbone LLMs}

We primarily experiment with three backbone LLMs that span different model scales and architectures:

\begin{itemize}
    \item \textbf{Qwen3-30B} (Qwen3-30B-A3B-Thinking-2507~\citep{qwen3technicalreport}), which serves as the strongest model in our main experimental setup. It adopts a Mixture-of-Experts (MoE) architecture with 30B total parameters and 3B activated parameters per token. This model delivers competitive reasoning capabilities while maintaining reasonable inference efficiency, making it our primary choice for complex trajectory planning tasks.
    \item \textbf{Qwen3-4B} (Qwen3-4B-Instruct-2507), a lightweight dense model with 4B parameters. It serves as an efficient and cost-effective variant. Despite its smaller size, it retains strong instruction-following and generalization abilities, allowing us to test the scalability and robustness of our method under constrained computational budgets.
    \item \textbf{LLaMA} (Llama-3.1-8B-Instruct)~\citep{LLaMA}, an 8B dense model from the Llama family. We include it to assess cross-family generalization and ensure that our approach is not overfit to the Qwen architecture series.
\end{itemize}

In addition to these primary models, we also conduct supplementary experiments on several auxiliary models to further verify the effectiveness and transferability of our method across different model families, scales, and training paradigms:

\begin{itemize}
    \item \textbf{GLM-Z1} (GLM-Z1-32B-0414)~\citep{GLM-Z1}, a 32B MoE model from the GLM family, is included to validate that our findings generalize beyond the Qwen architecture.
    \item \textbf{Qwen3.6-27B} (Qwen3.6-27B), a dense 27B model, represents the strongest model we can practically deploy within our hardware budget, serving as a powerful reference for assessing performance on more capable dense architectures.
    \item \textbf{Qwen3.6-35B-A3B} (Qwen3.6-35B-A3B), a Mixture-of-Experts variant with 35B total parameters and 3B activated per token, offers a favorable trade-off between capability and inference efficiency—delivering slightly lower performance than Qwen3.6-27B but with substantially faster inference.
    \item \textbf{Qwen3.5-9B} (Qwen3.5-9B), a mid-sized dense model from the latest Qwen generation, allows us to examine the performance scaling between our lightweight and large primary models.
    \item \textbf{TableGPT-R1}~\citep{TableGPT-R1}, a table-specialized model, is used to explore the potential of domain-specific LLMs in our trajectory planning framework.
\end{itemize}

Due to computational constraints, for all these auxiliary models, we run fewer experiments focusing only on key configurations to confirm that our main findings remain consistent beyond the primary model set, rather than conducting exhaustive hyperparameter sweeps.

\subsection{Default Configuration}

Unless otherwise specified, each query is allocated a budget of 5 or 10 interaction steps. During refinement, the agent adopts a ``select 1 out of 5 tables'' strategy (5s1). For row retrieval, each grep pattern returns at most one row and the total number of returned rows is capped at five. We denote these settings as \ours{}(5step-5s1-small) and \ours{}(10step-5s1-small), abbreviated as \ours{}(5-step) and \ours{}(10-step), respectively.

\subsection{Hardware and Deployment}

The embedding and reranker models are deployed using Transformers (v4.57.6, Python 3.9), and LLM inference is primarily powered by SGLang (v0.5.9, Python 3.10). For certain models, we additionally employ vLLM (v0.17.1, Python 3.11) as an alternative deployment backend, mainly because SGLang exhibits compatibility issues with these specific models, including deployment failures, runtime bugs, or unsupported features. All models are accessed via an OpenAI-compatible client interface. To ensure reproducibility, we set the sampling temperature to 0. In addition, the \texttt{enable-deterministic-inference} flag is enabled in SGLang. This setup is generally stable, though some randomness remains. It is worth noting that vLLM exhibits stronger randomness compared to SGLang; therefore, results from vLLM-deployed models are included only as supplementary references for comparison rather than as primary findings.

The experiments are conducted on two dedicated servers. The main server is equipped with dual Intel Xeon Gold 5218 CPUs @ 2.30GHz (16 cores each), 376~GB of RAM, and runs Ubuntu 18.04.6 LTS. We utilize one NVIDIA RTX A6000 (48~GB) for embedding and reranking, with driver version 525.147.05 and CUDA 12.4.

The separate inference server, dedicated to SGLang-based and vllm-based LLM deployment, features an AMD EPYC 7542 32-Core Processor, 1.0~TB of RAM, and Ubuntu 22.04.3 LTS, with driver version 550.90.07 and CUDA 12.4. On this server, we deploy LLMs using four NVIDIA RTX 4090 GPUs (24~GB each).

\paragraph{Reproducibility References.}
Readers unfamiliar with the frameworks used in this work are referred to the official documentation and tutorials available from the Hugging Face, vLLM and SGLang project homepages. A complete list of dependencies with version numbers is provided in the accompanying \texttt{requirements.txt} file.

\section{Hyperparameters}
\label{sec:hyperparameters}

We omit endpoint URLs, API keys, and filesystem paths.

\subsection{Shared Settings}

\begin{itemize}
    \item \textbf{Table serialization:} Specifies the output format. Available options: Markdown (\texttt{table\_format=markdown}), HTML (\texttt{table\_format=html}), CSV (\texttt{table\_format=csv}), or Mixed (\texttt{table\_format=mixed}) (sometimes written as \texttt{Default} in implementations, meaning to use the format defined in the table metadata file, which is equivalent to Mixed).
    \item \textbf{Query field:} Final processed query (as TCR-Bench introduces several variants of the original question/claim) (\texttt{content\_type=question}).
    \item \textbf{LLM backbone:} See the detailed model list in the ``Additional Experimental Details'' section.
\end{itemize}

\subsection{\ours{}}

Unless otherwise specified, \ours{} adopts the \textit{5s1-small} configuration used throughout the main experiments.

\begin{table*}[t]
\centering
\small
\begin{tabular}{lc}
\toprule
Hyperparameter & Value \\
\midrule
Max LLM steps per query (\texttt{max\_steps}) & 5 \\
Candidate tables shown per refine (\texttt{n\_candidates}) & 5 \\
Max tables selectable per refine (\texttt{max\_select\_tables}) & 1 \\
Table head preview length (\texttt{preview\_chars}) & 800 \\
Max matched rows per candidate (\texttt{max\_matched\_rows}) & 5 \\
Max matched rows per pattern (\texttt{max\_rows\_per\_pattern}) & 1 \\
Matched-row truncation length (\texttt{max\_line\_chars}) & 400 \\
Fallback tables for QA (\texttt{top\_k}) & 1 \\
\bottomrule
\end{tabular}
\caption{Default \ours{} (5s1-small) configuration.}
\label{tab:tableseek-default}
\end{table*}

Here, \texttt{max\_steps} counts all LLM calls for a query, including initial pattern generation, refine rounds, and final QA. At least one step is always reserved for QA. During refinement, the agent inspects the top \texttt{n\_candidates} tables ranked by pattern matches, each truncated to a head of \texttt{preview\_chars} characters plus up to \texttt{max\_matched\_rows} matched rows. Each pattern contributes at most \texttt{max\_rows\_per\_pattern} rows, and individual rows are truncated to \texttt{max\_line\_chars} characters. If refinement yields no selection, up to \texttt{top\_k} fallback tables are passed to QA.

\ours{} supports two task types: \texttt{tableqa} (TCR-Bench) and \texttt{fact\_verification} (TARGET-TabFact).

Different refine strategies modify only the candidate pool size and selection behavior.

\begin{table*}[t]
\centering
\small
\begin{tabular}{lcc}
\toprule
Configuration & \texttt{n\_candidates} & \texttt{max\_select\_tables} \\
\midrule
5s1 & 5 & 1 \\
5s3 & 5 & 3 \\
10s1 & 10 & 1 \\
10s3 & 10 & 3 \\
10s5 & 10 & 5 \\
r3 & 3 & 1 \\
r5 & 5 & 1 \\
r10 & 10 & 1 \\
\bottomrule
\end{tabular}
\caption{Refine configurations explored in this work.}
\label{tab:tableseek-refine-configs}
\end{table*}

For select-based strategies (e.g., 5s1 and 10s1), the agent explicitly selects up to \texttt{max\_select\_tables} tables from the candidate pool. For rerank-based variants (r$n$), the same candidate pool size is used, but candidates are reranked rather than explicitly selected, and the highest-ranked table is forwarded to the next iteration.

All stages of \ours{}, including pattern generation, refinement, TableQA, and TabFact verification, as well as direct QA, employ greedy decoding with a temperature of $0$. The decoding parameters are summarized in Table~\ref{tab:decoding-params}. Other hyperparameters related to API calls (e.g., timeout and retry configurations) are not expected to affect the generated outputs, as all requests completed successfully without triggering any fallback mechanisms. In practice, any remaining randomness primarily stems from the model deployment environment rather than the decoding strategy itself, and thus these settings have negligible influence on the final results.

\begin{table}[t]
\centering
\small
\begin{tabular}{lc}
\toprule
Parameter & Value \\
\midrule
\texttt{temperature} & $0.0$ \\
\texttt{max\_tokens} & $8192$ \\
\bottomrule
\end{tabular}
\caption{Decoding parameters used across all experiments.}
\label{tab:decoding-params}
\end{table}

\subsection{Direct QA}

Direct QA evaluates answer generation given gold tables or previously retrieved tables, without \ours{} retrieval. The table source and the number of retrieved tables are configured as shown in Table~\ref{tab:qa-hparams}. When \texttt{table\_source} is set to a specific retrieval method (e.g., \texttt{dense} or \texttt{rerank}), the top-$k$ tables from that method's retrieval results are used as input to QA, where $k$ corresponds to the \texttt{top\_k} hyperparameter.

\begin{table}[t]
\centering
\small
\begin{tabular}{lc}
\toprule
Hyperparameter & Value \\
\midrule
Table source (\texttt{table\_source}) & \texttt{gold} \\
Tables used for answering (\texttt{top\_k}) & 1 \\
\bottomrule
\end{tabular}
\caption{Direct QA hyperparameters.}
\label{tab:qa-hparams}
\end{table}

\subsection{Dense Retrieval}

Dense retrieval embeds each serialized table once and ranks tables according to cosine similarity with the query embedding. For the embedding models, we adopt the official pooling method provided by each model. We do not impose a hard cutoff on input length, instead using each model's maximum context window (or no limit when applicable). The detailed list of models is deferred to the ``Additional Experimental Details'' section.

\begin{table}[t]
\centering
\small
\begin{tabular}{lc}
\toprule
Hyperparameter & Value \\
\midrule
Retrieved tables (\texttt{top\_k}) & 5 \\
Similarity & Cosine \\
Embedding normalization & $\ell_2$ \\
\bottomrule
\end{tabular}
\caption{Dense retrieval hyperparameters.}
\label{tab:dense-hparams}
\end{table}

\subsection{ColBERT}

ColBERT employs a late interaction mechanism between query and document token representations. Queries and documents are encoded independently into token-level embeddings. During retrieval, the relevance score is computed as the sum of maximum cosine similarities between each query token embedding and the most similar document token embedding. Documents and queries are encoded with a batch size of $16$ each.

\subsection{Reranking}

For the reranker models, we adopt the official scoring method provided by each model. We do not impose a hard cutoff on input length, instead using each model's maximum context window (or no limit when applicable). The detailed list of models is deferred to the ``Additional Experimental Details'' section.

\begin{table}[t]
\centering
\small
\begin{tabular}{lc}
\toprule
Hyperparameter & Value \\
\midrule
Dense candidates as input & top-3/top-5 \\
Reranked tables (\texttt{top\_k}) & 3/5 \\
\bottomrule
\end{tabular}
\caption{Reranking hyperparameters.}
\label{tab:rerank-hparams}
\end{table}

The dense stage therefore retrieves three/five tables, after which the reranker rescales that subset and keeps the top three/five.

Unless otherwise noted, all retrieval baselines, including Dense, Rerank and SPLADE, are evaluated using their original released checkpoints and standard inference procedures. We do not apply model quantization, additional context-length restrictions beyond those imposed by the models themselves, or other efficiency-oriented modifications.

\subsection{LLM Serving Configuration}

All \ours{}-related LLM calls are served through SGLang and vLLM. Table~\ref{tab:sglang-config} and Table~\ref{tab:vllm-config} summarize the serving configurations for models deployed with SGLang and vLLM respectively.

For SGLang deployment in Table~\ref{tab:sglang-config}, TP denotes tensor parallelism size, Attn. backend refers to the attention backend, Ctx. len. is the context length, Max pref. and Chunk. pref. stand for maximum prefill tokens and chunked prefill size respectively, and Mem. frac. indicates the memory fraction.

For vLLM deployment in Table~\ref{tab:vllm-config}, TP denotes tensor parallelism size, Mem. util. refers to GPU memory utilization, and Max len.

Unless otherwise specified, the context length is set to 40,960. Three exceptions apply: GLM-Z1-32B-0414 uses its native maximum of 32,768, and Qwen3-30B-A3B-Thinking-2507 is set to 65,536 to accommodate the 5s1-full evaluation setting. For Qwen3.6-35B-A3B, we set the context length to 32,768 due to deployment constraints; this is sufficient for the vast majority of our tasks, as 40,960 is adopted as a conservative upper bound and only the 5s1-full setting occasionally exceeds 40,960.

\begin{table*}[t]
\centering
\small
\begin{tabular}{lccccccc}
\toprule
Model & TP & Attn. backend & Ctx. len. & Max pref. & Chunk. pref. & Mem. frac. & Other \\
\midrule
Qwen3-30B-A3B-Thinking-2507 & 4 & FlashInfer & 65,536 & 65,536 & 4,096 & 0.8 & - \\
Qwen3-4B-Instruct-2507 & 2 & FlashInfer & 40,960 & 40,960 & 4,096 & 0.85 & - \\
GLM-Z1-32B-0414 & 4 & FlashInfer & 32,768 & 32,768 & 4,096 & 0.8 & - \\
Llama-3.1-8B-Instruct & 2 & Default & 40,960 & 40,960 & 4,096 & 0.8 & - \\
\bottomrule
\end{tabular}
\caption{SGLang serving configuration for different models.}
\label{tab:sglang-config}
\end{table*}

\begin{table*}[t]
\centering
\small
\begin{tabular}{lcccc}
\toprule
Model & TP & Mem. util. & Max len. & Other \\
\midrule
Qwen3.5-9B & 4 & 0.9 & 40,960 & Default \\
Qwen3.6-35B-A3B & 4 & 0.85 & 32,768 & Default \\
Qwen3.6-27B & 4 & 0.8 & 40,960 & Default \\
TableGPT-R1 & 4 & 0.85 & 40,960 & Default \\
\bottomrule
\end{tabular}
\caption{vLLM serving configuration for different models.}
\label{tab:vllm-config}
\end{table*}

\section{Prompts Used in Our Pipeline}
\label{sec:prompt}
\newtcolorbox{promptbox}[1]{
    colback=white,
    colframe=black,
    coltitle=black,
    colbacktitle=white,
    fonttitle=\bfseries,
    title=#1,
    boxrule=0.8pt,
    arc=0pt,
    left=5pt,
    right=5pt,
    top=5pt,
    bottom=5pt,
    standard jigsaw,
    toptitle=2pt,
    bottomtitle=2pt,
    width=0.9\textwidth
}

\begin{table*}[!tbp]
\centering
\begin{promptbox}{Task Instruction for TCR-Bench}
\small
\texttt{Given a TableQA query, retrieve a relevant table that can answer the query. Note that the retrieved table should contain sufficient information to provide an answer, rather than resulting in an empty answer.}
\end{promptbox}
\caption{Task instruction for TCR-Bench.}
\label{tab:retrieval-prompt-tcr}
\end{table*}

\begin{table*}[!tbp]
\centering
\begin{promptbox}{Task Instruction for TARGET-TabFact}
\small
\texttt{Given a claim about tabular data, retrieve a relevant table whose schema and content can be used to verify the claim. Focus on tables whose column names and row structure are relevant to the claim, even if an exact answer cannot be directly read off.}
\end{promptbox}
\caption{Task instruction for TARGET-TabFact.}
\label{tab:retrieval-prompt-tabfact}
\end{table*}

\begin{table*}[!tbp]
\centering
\begin{promptbox}{GREP System Prompt}
\small
\textbf{System Prompt:} \\
You are a table retrieval assistant. Your job is to identify which tables are relevant to a user query by generating regex patterns (like grep) to match against table text strings. \\

\smallskip
You will be given: \\
1. A natural language query \\
2. A list of table keys (identifiers) \\

\smallskip
Your task: \\
- Analyse the query carefully \\
- Generate a list of Python-compatible regex patterns (case-insensitive) that, when matched against a table's text representation, would indicate that table is relevant \\
- Patterns should target: column names, entity names, key terms, or values likely present in a relevant table \\
- Avoid overly generic patterns (e.g. single common words) \\
- Prefer specific noun phrases, entity names, numeric identifiers, or domain-specific column names \\

\smallskip
Output format — respond with a JSON object (no markdown fences, no extra text): \\
\verb|{| \\
\verb|  "patterns": ["pattern1", "pattern2", ...]| \\
\verb|}| \\

\smallskip
The patterns will be applied with \texttt{re.search(..., table\_text, re.IGNORECASE)}. \\
Generate between 3 and 10 patterns. \\

\medskip
\hrule
\medskip

\textbf{User Prompt:} \\
Task context: \texttt{\{task\_hint\}} \\

Query: \texttt{\{query\}} \\

Generate regex patterns to grep relevant tables.
\end{promptbox}
\caption{System and user prompt templates for GREP-based table retrieval in TCR-Bench.}
\label{tab:grep-prompt}
\end{table*}

\begin{table*}[!tbp]
\centering
\begin{promptbox}{Refine Prompt}
\small
\textbf{System Prompt:} \\
You are an expert table retrieval assistant helping to iteratively refine a search based on table previews.
You will be given:
1. The original query
2. The top candidate table previews, labeled with anonymous IDs like ``Table\_0'', ``Table\_1'', etc.
   Each preview may include:
   - the beginning of the table (truncated if large)
   - an optional section ``Rows matched by current grep patterns'' listing rows/lines that matched the regex patterns
3. The current regex patterns used

\smallskip
\textbf{Heuristic Guidelines for Decision Making:}
\begin{itemize}
\item \textbf{Select (High Confidence):} If one or more candidate tables clearly contain the data needed to answer the query, select them immediately. \texttt{\{select\_guidance\}}
\item \textbf{Regenerate by Value Search (Schema Collision):} If the top K candidate tables share very similar structures/schemas, but none of their visible values match the query's target entity, DO NOT select randomly. Instead, generate new regex patterns focusing on specific values, keywords, or entities from the original query.
\item \textbf{Regenerate by Schema Shift (Irrelevant Results):} If all candidate tables are completely irrelevant, generate new regex patterns exploring alternative schemas, synonyms, or different table structures.
\item \textbf{Regenerate by Broadening (Narrow Misses):} If the tables are in the right domain but the format is slightly off, generate broader or fuzzy regex patterns.
\end{itemize}

\smallskip
\textbf{Output Format:} \\
Respond with a JSON object (no markdown fences, no extra text). \\
\texttt{\{output\_select\_hint\}} \\
\verb|{ "action": "select", "selected_ids": |\texttt{\{example\_ids\}}\verb| }| \\
or \\
\verb|{ "action": "regenerate", "patterns": ["pattern1", ...] }| \\

\smallskip
Be decisive and strategic. Analyze the gap between the query and the candidates to choose the best regeneration strategy.

\medskip
\hrule
\medskip

\textbf{User Prompt:} \\
Task context: \texttt{\{task\_hint\}} \\

Query: \texttt{\{query\}} \\

Current patterns: \texttt{\{current\_patterns\_json\}} \\

Top candidate table previews: \\
\texttt{\{table\_previews\}} \\

\texttt{\{select\_instruction\}}
\end{promptbox}
\caption{System and user prompt templates for the multi-round refinement retriever. The \texttt{\{select\_guidance\}}, \texttt{\{output\_select\_hint\}}, \texttt{\{example\_ids\}}, and \texttt{\{select\_instruction\}} placeholders are instantiated conditionally based on \texttt{max\_select\_tables} (1 vs. $>$1). The \texttt{\{table\_previews\}} placeholder expands to a sequence of anonymous table IDs with their previews and any rows matched by the current grep patterns.}
\label{tab:refine-prompt}
\end{table*}

\begin{table*}[!tbp]
\centering
\begin{promptbox}{Refine Prompt for Rerank Variant}
\small
\textbf{System Prompt:} \\
You are an expert table retrieval assistant helping to iteratively refine a search based on table previews.
You will be given:
1. The original query
2. The top candidate table previews, labeled with anonymous IDs like ``Table\_0'', ``Table\_1'', etc.
   Each preview may include:
   - the beginning of the table (truncated if large)
   - an optional section ``Rows matched by current grep patterns'' listing rows/lines that matched the regex patterns (grep is row-oriented even though retrieval ranks whole tables)
3. The current regex patterns used

\smallskip
\textbf{Heuristic Guidelines for Decision Making:}
\begin{itemize}
\item \textbf{Rerank (Default):} Re-order ALL candidate tables from most to least likely to answer the query. Put the table most likely to contain the answer first. Include every candidate anonymous ID exactly once.
\item \textbf{Regenerate by Value Search (Schema Collision):} If the top K candidate tables share very similar structures/schemas (e.g., same column headers), but none of their visible values match the query's target entity, DO NOT rerank randomly. Instead, generate new regex patterns focusing on \textbf{specific values, keywords, or entities} from the original query to filter out the correct table.
\item \textbf{Regenerate by Schema Shift (Irrelevant Results):} If all candidate tables are completely irrelevant (wrong domain, wrong format, or totally different context), your previous schema-based regex has failed. Generate new regex patterns exploring \textbf{alternative schemas, synonyms, or different table structures} that might exist in the database.
\item \textbf{Regenerate by Broadening (Narrow Misses):} If the tables are in the right domain but the format is slightly off, generate broader or fuzzy regex patterns to catch variations.
\end{itemize}

\smallskip
\textbf{Output Format:} \\
Respond with a JSON object (no markdown fences, no extra text). \\
If reranking candidates (include every candidate ID exactly once, best first): \\
\verb|{ "action": "rerank", "ranked_ids": ["Table_2", "Table_0", "Table_1"] }| \\
or \\
\verb|{ "action": "regenerate", "patterns": ["pattern1", "pattern2", "pattern3"] }| \\

\smallskip
Be decisive and strategic. Analyze the gap between the query and the candidates to choose the best regeneration strategy.

\medskip
\hrule
\medskip

\textbf{User Prompt:} \\
Task context: \texttt{\{task\_hint\}} \\

Query: \texttt{\{query\}} \\

Current patterns: \texttt{\{current\_patterns\_json\}} \\

Top candidate table previews: \\
\texttt{\{table\_previews\}} \\

Decide: rerank all candidate tables from most to least relevant (by anonymous IDs), or regenerate patterns if the candidates are all wrong.
\end{promptbox}
\caption{System and user prompt templates for the multi-round refinement retriever for rerank variant (ChunkR). The \texttt{\{task\_hint\}} placeholder is instantiated from \texttt{QUERY\_PREFIX\_MAP[task\_type]} (e.g., \texttt{fetaqa} or \texttt{text2sql}). The \texttt{\{table\_previews\}} placeholder expands to a sequence of anonymous table IDs with their previews, formatted as \texttt{[Table\_0] ... preview ... [Table\_1] ... preview ...}. Unlike the selection-based variant in Table~\ref{tab:refine-prompt}, this version outputs a \texttt{rerank} action that reorders all candidates rather than selecting a subset.}
\label{tab:refine-prompt-rerank}
\end{table*}

\begin{table*}[!tbp]
\centering
\begin{promptbox}{QA Prompt for TCR-Bench}
\small
\textbf{System Prompt:} \\
\#\#\# Requirements: \\
Please read the following table and then answer the questions based on the table. Organize your answers into a list of strings, with each element being an answer item (The number of answer items is less than 11). If the question includes a special requirement, such as outputting a dictionary, please fulfill that specific request. Otherwise, always output a list of strings, even if there is only one answer item. Please place your answers between the \verb|```| and \verb|```|. \\

\smallskip
\#\#\# Notes: \\
The table may include non-standard formats for numbers or dates. For numbers, formats may include Roman numerals, English words, or scientific notation. Whenever possible, please convert these into Arabic numerals in your response. \\
Additionally, some columns are derived through mathematical operations. For example, \texttt{total\_add\_10} (other operations or variants may also exist) indicates that the values in this column are obtained by adding 10 to the original values. You should return the original values (i.e., subtract 10 from the current values). Furthermore, for any values ending with ``k'' or ``K'', convert them into standard Arabic numerals. Please convert all dates to the format \%Y-\%m-\%d <other time elements> (in Arabic numerals) in your response. \\

\smallskip
Additionally, even if there are duplicate answer items, you need to output all of them. \\

\smallskip
\# Example Output Format: \\
\verb|```| \\
\verb|['answer_item_1', 'answer_item_2', ..., 'answer_item_n']| \\
\verb|```| \\

\smallskip
Please do not generate any text after outputting the final answer.

\medskip
\hrule
\medskip

\textbf{User Prompt:} \\
table information: \\
\texttt{\{table\_infos\}} \\

Question: \texttt{\{query\}} \\
Answer:
\end{promptbox}
\caption{System and user prompt templates for the final QA stage in \ours{} on TCR-Bench. The \texttt{\{table\_infos\}} placeholder is instantiated with concatenated retrieved tables in the format \texttt{[Table: \{key\}]}\textbackslash n\texttt{\{table\_str\}}.}
\label{tab:tcr-prompt}
\end{table*}

\begin{table*}[!tbp]
\centering
\begin{promptbox}{QA Prompt for TARGET-TabFact}
\small
\textbf{System Prompt:} \\
Determine whether the given statement is supported by the retrieved table. \\

\smallskip
Requirements: \\
- Use only information from the table. \\
- If the statement is fully supported by the table, output `true`. \\
- If the statement is contradicted by the table, output `false`. \\
- If the retrieved table does not contain the columns or entities needed to verify the statement (i.e., schema mismatch), output `mismatch`. \\
- Do not provide any explanation. \\
- Output only one word: `true`, `false`, or `mismatch`. \\

\medskip
\hrule
\medskip

\textbf{User Prompt:} \\
table information: \\
\texttt{\{table\_infos\}} \\

Statement: \texttt{\{statement\}} \\
Answer:
\end{promptbox}
\caption{System and user prompt templates for the fact verification (TabFact) stage in \ours{} on TARGET-TabFact. The \texttt{\{table\_infos\}} placeholder is instantiated with concatenated retrieved tables in the format \texttt{[Table: \{key\}]}\textbackslash n\texttt{\{table\_str\}}.}
\label{tab:tabfact-prompt}
\end{table*}

\begin{table*}[!tbp]
\centering
\begin{promptbox}{Standalone QA Prompt for TCR-Bench}
\small
\textbf{System Prompt:} \\
\#\#\# Requirements: \\
Please read the following table and then answer the questions based on the table. Organize your answers into a list of strings, with each element being an answer item (The number of answer items is less than 11). If the question includes a special requirement, such as outputting a dictionary, please fulfill that specific request. Otherwise, always output a list of strings, even if there is only one answer item. Please place your answers between the \verb|```| and \verb|```|. \\

\smallskip
\#\#\# Notes: \\
The table may include non-standard formats for numbers or dates. For numbers, formats may include Roman numerals, English words, or scientific notation. Whenever possible, please convert these into Arabic numerals in your response. \\
Additionally, some columns are derived through mathematical operations. For example, \texttt{total\_add\_10} indicates that the values in this column are obtained by adding 10 to the original values. You should return the original values (i.e., subtract 10 from the current values). Furthermore, for any values ending with ``k'' or ``K'', convert them into standard Arabic numerals. Please convert all dates to the format \%Y-\%m-\%d <other time elements> (in Arabic numerals) in your response. \\

\smallskip
Additionally, even if there are duplicate answer items, you need to output all of them. \\

\smallskip
\# Example Output Format: \\
\verb|```| \\
\verb|['answer_item_1', 'answer_item_2', ..., 'answer_item_n']| \\
\verb|```| \\

\smallskip
Please do not generate any text after outputting the final answer.

\medskip
\hrule
\medskip

\textbf{User Prompt:} \\
table information: \\
\texttt{\{table\_infos\}} \\

Question: \texttt{\{query\}} \\
Answer:
\end{promptbox}
\caption{System and user prompt templates for the standalone QA stage used in the dense-retrieval-then-rerank-then-QA pipeline. This template is similar to \ours{}'s QA prompt (Table~\ref{tab:tcr-prompt}) but omits the ``other operations or variants'' clause in the mathematical operations note.}
\label{tab:standalone-tcr-prompt}
\end{table*}

\begin{table*}[!tbp]
\centering
\begin{promptbox}{Standalone QA Prompt for TARGET-TabFact}
\small
\textbf{System Prompt:} \\
Determine whether the given statement is supported by the table. \\

\smallskip
Requirements: \\
- Use only information from the table. \\
- If the statement is fully supported by the table, output `true`. \\
- If the statement is contradicted by the table or cannot be verified from the table, output `false`. \\
- Do not provide any explanation. \\
- Output only one word: `true` or `false`. \\

\medskip
\hrule
\medskip

\textbf{User Prompt:} \\
table information: \\
\texttt{\{table\_infos\}} \\

Question: \texttt{\{query\}} \\
Answer:
\end{promptbox}
\caption{System and user prompt templates for the standalone fact verification (TabFact) stage used in the dense-retrieval-then-rerank-then-QA pipeline. Unlike \ours{}'s TabFact prompt (Table~\ref{tab:tabfact-prompt}), this version (1) does not include the \texttt{mismatch} output option, (2) treats unverifiable cases as \texttt{false}, and (3) uses ``Question:'' as the user label instead of ``Statement:''.}
\label{tab:standalone-tabfact-prompt}
\end{table*}
The prompts we employed are relatively complex. For TCR-Bench~\citep{TCR-Bench} and TARGET-TabFact~\citep{TARGET, TabFact}, we also used different prefixes and final QA prompts.

For the prompts related to dense retrieval and embedding, we adopt those from TCR-Bench.

Table~\ref{tab:standalone-tcr-prompt} and Table~\ref{tab:standalone-tabfact-prompt} present the standard QA prompts, which are primarily used in the Dense-Rerank-QA pipeline.

Table~\ref{tab:tabfact-prompt} and Table~\ref{tab:retrieval-prompt-tabfact} show the prompt prefixes for the two tasks under \ours{}, respectively. These prefixes are used during pattern generation and refinement to define the task boundaries.

Table~\ref{tab:grep-prompt} provides the prompt used for pattern generation in \ours{}.

Table~\ref{tab:refine-prompt} shows the prompt used for refinement in \ours{} when the action is `select` (i.e., picking 1..N tables). This prompt is relatively complicated, as it varies depending on the specific $\texttt{max\_select\_tables}$ setting (1 vs. $>$1), and it also includes substantial content that needs to be dynamically inserted. For the rerank variant where the action is `rerank` (reordering all candidates), we use a different version of this prompt tailored for that action, as shown in Table~\ref{tab:refine-prompt-rerank}.

Finally, Table~\ref{tab:tcr-prompt} and Table~\ref{tab:tabfact-prompt} present the prompts used for the final QA stage of \ours{}. Since these prompts incorporate fallback conditions, they differ from the standard QA prompts shown in Table~\ref{tab:standalone-tcr-prompt} and Table~\ref{tab:standalone-tabfact-prompt}.

\section{Additional Backbone Analysis}
\label{sec:appendix_backbone}

This section provides additional backbone comparison analysis beyond the main results reported in the paper, covering observations on mixed-format data. We include supplementary experimental insights across different backbone models and examine how backbone capability and search budget influence the performance of \ours{} in greater detail.
\begin{table*}[!tbp]
\centering
\begin{tabular}{llccccc}
\toprule
Method & Config & Format & Recall & F1 (Qwen3-30B) & F1 (Qwen3-4B) & F1 (LLaMA) \\
\midrule
Dense & E4B & CSV & 0.158 & 0.116 & 0.054 & 0.079 \\
Dense & E4B & Mixed & 0.172 & 0.118 & 0.064 & 0.088 \\
Dense & E4B & HTML & 0.134 & 0.122 & 0.048 & 0.056 \\
Dense & E4B & Markdown & 0.187 & 0.134 & 0.077 & 0.109 \\
\midrule
Rerank (top-3) & E4B+R4B & CSV & 0.354 & 0.284 & 0.102 & 0.173 \\
Rerank (top-3) & E4B+R4B & Mixed & 0.359 & 0.269 & 0.127 & 0.172 \\
Rerank (top-3) & E4B+R4B & HTML & 0.373 & 0.279 & 0.140 & 0.165 \\
Rerank (top-3) & E4B+R4B & Markdown & 0.407 & 0.314 & 0.137 & 0.173 \\
\midrule
Rerank (top-5) & E4B+R4B & CSV & 0.445 & 0.356 & 0.141 & 0.199 \\
Rerank (top-5) & E4B+R4B & Mixed & 0.421 & 0.317 & 0.137 & 0.198 \\
Rerank (top-5) & E4B+R4B & HTML & 0.459 & 0.353 & 0.171 & 0.205 \\
Rerank (top-5) & E4B+R4B & Markdown & 0.488 & 0.362 & 0.167 & 0.189 \\
\midrule
\ours{} (5-step) & Qwen3-4B & CSV & 0.172 & - & 0.075 & - \\
\ours{} (5-step) & Qwen3-4B & Mixed & 0.163 & - & 0.068 & - \\
\ours{} (5-step) & Qwen3-4B & HTML & 0.177 & - & 0.075 & - \\
\ours{} (5-step) & Qwen3-4B & Markdown & 0.182 & - & 0.087 & - \\
\midrule
\ours{} (5-step) & LLaMA & CSV & 0.206 & - & - & 0.096 \\
\ours{} (5-step) & LLaMA & Mixed & 0.182 & - & - & 0.067 \\
\ours{} (5-step) & LLaMA & HTML & 0.234 & - & - & 0.105 \\
\ours{} (5-step) & LLaMA & Markdown & 0.187 & - & - & 0.096 \\
\midrule
\ours{} (5-step) & Qwen3-30B & CSV & 0.411 & 0.283 & - & - \\
\ours{} (5-step) & Qwen3-30B & Mixed & 0.416 & 0.284 & - & - \\
\ours{} (5-step) & Qwen3-30B & HTML & 0.392 & 0.292 & - & - \\
\ours{} (5-step) & Qwen3-30B & Markdown & 0.416 & 0.318 & - & - \\
\midrule
\ours{} (10-step) & Qwen3-4B & CSV & 0.196 & - & 0.073 & - \\
\ours{} (10-step) & Qwen3-4B & Mixed & 0.196 & - & 0.071 & - \\
\ours{} (10-step) & Qwen3-4B & HTML & 0.163 & - & 0.069 & - \\
\ours{} (10-step) & Qwen3-4B & Markdown & 0.211 & - & 0.095 & - \\
\midrule
\ours{} (10-step) & LLaMA & CSV & 0.220 & - & - & 0.110 \\
\ours{} (10-step) & LLaMA & Mixed & 0.196 & - & - & 0.070 \\
\ours{} (10-step) & LLaMA & HTML & 0.249 & - & - & 0.120 \\
\ours{} (10-step) & LLaMA & Markdown & 0.201 & - & - & 0.091 \\
\midrule
\ours{} (10-step) & Qwen3-30B & CSV & 0.493 & 0.346 & - & - \\
\ours{} (10-step) & Qwen3-30B & Mixed & 0.507 & 0.361 & - & - \\
\ours{} (10-step) & Qwen3-30B & HTML & 0.498 & 0.364 & - & - \\
\ours{} (10-step) & Qwen3-30B & Markdown & 0.498 & 0.364 & - & - \\
\bottomrule
\end{tabular}
\caption{Full backbone comparison on TCR-Bench with Qwen3-30B, Qwen3-4B, and LLaMA. Recall is R@1 for Dense/Rerank and R@all for \ours{}.}
\label{tab:appendix_tcr_full}
\end{table*}

\begin{table*}[!tbp]
\centering
\begin{tabular}{llccc}
\toprule
Method & Config & Recall & F1 \\
\midrule
Oracle & GLM-Z1 & 1.000 & 0.501 \\
Dense & E4B+GLM-Z1 & 0.172 & 0.099 \\
Rerank (top-3) & E4B+R4B+GLM-Z1 & 0.359 & 0.197 \\
Rerank (top-5) & E4B+R4B+GLM-Z1 & 0.421 & 0.264 \\
\ours{} (5-step-5s1-small) & GLM-Z1 & 0.258 & 0.170 \\
\ours{} (10-step-5s1-small) & GLM-Z1 & 0.306 & 0.180 \\
\midrule
Oracle & Qwen3.5-9B & 1.000 & 0.564 \\
Dense & E4B+Qwen3.5-9B & 0.172 & 0.108 \\
Rerank (top-3) & E4B+R4B+Qwen3.5-9B & 0.359 & 0.210 \\
Rerank (top-5) & E4B+R4B+Qwen3.5-9B & 0.421 & 0.250 \\
\ours{} (5-step-5s1-small) & Qwen3.5-9B & 0.512 & 0.335 \\
\ours{} (10-step-5s1-small) & Qwen3.5-9B & 0.579 & 0.369 \\
\midrule
Oracle & Qwen3.6-35B-A3B & 1.000 & 0.765 \\
Dense & E4B+Qwen3.6-35B-A3B & 0.172 & 0.150 \\
Rerank (top-3) & E4B+R4B+Qwen3.6-35B-A3B & 0.359 & 0.261 \\
Rerank (top-5) & E4B+R4B+Qwen3.6-35B-A3B & 0.421 & 0.336 \\
\ours{} (5-step-5s1-small) & Qwen3.6-35B-A3B & 0.574 & 0.470 \\
\ours{} (10-step-5s1-small) & Qwen3.6-35B-A3B & 0.632 & 0.503 \\
\midrule
Oracle & Qwen3.6-27B & 1.000 & 0.769 \\
Dense & E4B+Qwen3.6-27B & 0.172 & 0.151 \\
Rerank (top-3) & E4B+R4B+Qwen3.6-27B & 0.359 & 0.323 \\
Rerank (top-5) & E4B+R4B+Qwen3.6-27B & 0.421 & 0.359 \\
\ours{} (5-step-5s1-small) & Qwen3.6-27B & 0.641 & 0.545 \\
\ours{} (10-step-5s1-small) & Qwen3.6-27B & 0.684 & 0.591 \\
\midrule
Oracle & TableGPT-R1 & 1.000 & 0.637 \\
Dense & E4B+TableGPT-R1 & 0.172 & 0.129 \\
Rerank (top-3) & E4B+R4B+TableGPT-R1 & 0.359 & 0.241 \\
Rerank (top-5) & E4B+R4B+TableGPT-R1 & 0.421 & 0.277 \\
\ours{} (5-step-5s1-small) & TableGPT-R1 & 0.359 & 0.233 \\
\ours{} (10-step-5s1-small) & TableGPT-R1 & 0.421 & 0.307 \\
\bottomrule
\end{tabular}
\caption{Supplementary backbone comparison on TCR-Bench with additional models beyond the three primary backbones. Recall is R@1 for Dense/Rerank and R@all for \ours{}.}
\label{tab:appendix_tcr_others}
\end{table*}

\begin{table*}[!tbp]
\centering
\begin{tabular}{lllccccc}
\toprule
Method & Split & Config & Retrieval & Acc & Acc-Sup & Acc-Ref \\
\midrule
Dense & test & Qwen3-EB+Qwen3-4B & 0.495 & 0.628 & 0.407 & 0.851 \\
Dense & test-1k & Qwen3-EB+Qwen3-4B & 0.498 & 0.610 & 0.405 & 0.823 \\
Dense & test & Stella+Qwen3-4B & 0.520 & 0.637 & 0.435 & 0.840 \\
Dense & test-1k & Stella+Qwen3-4B & 0.519 & 0.627 & 0.446 & 0.815 \\
Dense & test & Qwen3-EB+LLaMA & 0.495 & 0.611 & 0.308 & 0.918 \\
Dense & test-1k & Qwen3-EB+LLaMA & 0.498 & 0.598 & 0.297 & 0.910 \\
Dense & test & Stella+LLaMA & 0.520 & 0.619 & 0.327 & 0.913 \\
Dense & test-1k & Stella+LLaMA & 0.519 & 0.618 & 0.334 & 0.912 \\
Dense & test & Qwen3-EB+Qwen3-30B & 0.495 & 0.708 & 0.468 & 0.950 \\
Dense & test-1k & Qwen3-EB+Qwen3-30B & 0.498 & 0.701 & 0.456 & 0.955 \\
Dense & test & Stella+Qwen3-30B & 0.520 & 0.718 & 0.490 & 0.948 \\
Dense & test-1k & Stella+Qwen3-30B & 0.519 & 0.721 & 0.489 & 0.961 \\
\midrule
Rerank (top-3) & test & Qwen3-E4B+Qwen3-R4B+Qwen3-4B & 0.603 & 0.657 & 0.500 & 0.816 \\
Rerank (top-3) & test-1k & Qwen3-E4B+Qwen3-R4B+Qwen3-4B & 0.604 & 0.633 & 0.483 & 0.788 \\
Rerank (top-3) & test & Stella+Qwen3-R4B+Qwen3-4B & 0.612 & 0.658 & 0.510 & 0.808 \\
Rerank (top-3) & test-1k & Stella+Qwen3-R4B+Qwen3-4B & 0.623 & 0.643 & 0.501 & 0.790 \\
Rerank (top-3) & test & Qwen3-E4B+Qwen3-R4B+LLaMA & 0.603 & 0.639 & 0.376 & 0.905 \\
Rerank (top-3) & test-1k & Qwen3-E4B+Qwen3-R4B+LLaMA & 0.604 & 0.625 & 0.360 & 0.900 \\
Rerank (top-3) & test & Stella+Qwen3-R4B+LLaMA & 0.612 & 0.642 & 0.384 & 0.902 \\
Rerank (top-3) & test-1k & Stella+Qwen3-R4B+LLaMA & 0.623 & 0.629 & 0.369 & 0.898 \\
Rerank (top-3) & test & Qwen3-E4B+Qwen3-R4B+Qwen3-30B & 0.603 & 0.758 & 0.576 & 0.942 \\
Rerank (top-3) & test-1k & Qwen3-E4B+Qwen3-R4B+Qwen3-30B & 0.604 & 0.754 & 0.562 & 0.953 \\
Rerank (top-3) & test & Stella+Qwen3-R4B+Qwen3-30B & 0.612 & 0.760 & 0.583 & 0.938 \\
Rerank (top-3) & test-1k & Stella+Qwen3-R4B+Qwen3-30B & 0.623 & 0.752 & 0.562 & 0.949 \\
\midrule
Rerank (top-5) & test & Qwen3-E4B+Qwen3-R4B+Qwen3-4B & 0.641 & 0.668 & 0.534 & 0.803 \\
Rerank (top-5) & test-1k & Qwen3-E4B+Qwen3-R4B+Qwen3-4B & 0.635 & 0.638 & 0.513 & 0.768 \\
Rerank (top-5) & test & Stella+Qwen3-R4B+Qwen3-4B & 0.647 & 0.668 & 0.541 & 0.796 \\
Rerank (top-5) & test-1k & Stella+Qwen3-R4B+Qwen3-4B & 0.656 & 0.648 & 0.525 & 0.776 \\
Rerank (top-5) & test & Qwen3-E4B+Qwen3-R4B+LLaMA & 0.641 & 0.647 & 0.398 & 0.900 \\
Rerank (top-5) & test-1k & Qwen3-E4B+Qwen3-R4B+LLaMA & 0.635 & 0.636 & 0.387 & 0.894 \\
Rerank (top-5) & test & Stella+Qwen3-R4B+LLaMA & 0.647 & 0.649 & 0.405 & 0.896 \\
Rerank (top-5) & test-1k & Stella+Qwen3-R4B+LLaMA & 0.656 & 0.641 & 0.397 & 0.894 \\
Rerank (top-5) & test & Qwen3-E4B+Qwen3-R4B+Qwen3-30B & 0.641 & 0.775 & 0.615 & 0.936 \\
Rerank (top-5) & test-1k & Qwen3-E4B+Qwen3-R4B+Qwen3-30B & 0.635 & 0.771 & 0.599 & 0.949 \\
Rerank (top-5) & test & Stella+Qwen3-R4B+Qwen3-30B & 0.647 & 0.775 & 0.616 & 0.935 \\
Rerank (top-5) & test-1k & Stella+Qwen3-R4B+Qwen3-30B & 0.656 & 0.754 & 0.572 & 0.928 \\
\midrule
\ours{} (5-step) & test-1k & Qwen3-4B & 0.542 & 0.632 & 0.481 & 0.788 \\
\ours{} (5-step) & test-1k & LLaMA & 0.432 & 0.586 & 0.255 & 0.929 \\
\ours{} (5-step) & test-1k & Qwen3-30B & 0.617 & 0.754 & 0.566 & 0.949 \\
\ours{} (10-step) & test-1k & Qwen3-4B & 0.553 & 0.632 & 0.483 & 0.786 \\
\ours{} (10-step) & test-1k & LLaMA & 0.472 & 0.599 & 0.287 & 0.923 \\
\ours{} (10-step) & test-1k & Qwen3-30B & 0.648 & 0.764 & 0.589 & 0.945 \\
\bottomrule
\end{tabular}
\caption{Results on TARGET-TabFact-test and TARGET-TabFact-test-1k. Retrieval is R@1 for Dense/Rerank and R@all for \ours{}.}
\label{tab:main_tabfact_full}
\end{table*}

\begin{table*}[!tbp]
\centering
\begin{tabular}{lllccccc}
\toprule
Method & Split & Config & Retrieval & Acc & Acc-Sup & Acc-Ref \\
\midrule
Dense & valid & Qwen3-EB+Qwen3-4B & 0.492 & 0.633 & 0.415 & 0.857 \\
Dense & valid-1k & Qwen3-EB+Qwen3-4B & 0.515 & 0.610 & 0.395 & 0.814 \\
Dense & valid & Stella+Qwen3-4B & 0.516 & 0.639 & 0.434 & 0.849 \\
Dense & valid-1k & Stella+Qwen3-4B & 0.523 & 0.629 & 0.418 & 0.830 \\
Dense & valid & Qwen3-EB+LLaMA & 0.492 & 0.610 & 0.315 & 0.914 \\
Dense & valid-1k & Qwen3-EB+LLaMA & 0.515 & 0.610 & 0.303 & 0.902 \\
Dense & valid & Stella+LLaMA & 0.516 & 0.617 & 0.327 & 0.915 \\
Dense & valid-1k & Stella+LLaMA & 0.523 & 0.617 & 0.303 & 0.916 \\
Dense & valid & Qwen3-EB+Qwen3-30B & 0.492 & 0.705 & 0.464 & 0.952 \\
Dense & valid-1k & Qwen3-EB+Qwen3-30B & 0.515 & 0.709 & 0.453 & 0.953 \\
Dense & valid & Stella+Qwen3-30B & 0.516 & 0.713 & 0.484 & 0.948 \\
Dense & valid-1k & Stella+Qwen3-30B & 0.523 & 0.706 & 0.453 & 0.947 \\
\midrule
Rerank (top-3) & valid & Qwen3-E4B+Qwen3-R4B+Qwen3-4B & 0.600 & 0.663 & 0.502 & 0.827 \\
Rerank (top-3) & valid-1k & Qwen3-E4B+Qwen3-R4B+Qwen3-4B & 0.605 & 0.646 & 0.475 & 0.809 \\
Rerank (top-3) & valid & Stella+Qwen3-R4B+Qwen3-4B & 0.618 & 0.668 & 0.521 & 0.819 \\
Rerank (top-3) & valid-1k & Stella+Qwen3-R4B+Qwen3-4B & 0.621 & 0.643 & 0.486 & 0.793 \\
Rerank (top-3) & valid & Qwen3-E4B+Qwen3-R4B+LLaMA & 0.600 & 0.637 & 0.383 & 0.897 \\
Rerank (top-3) & valid-1k & Qwen3-E4B+Qwen3-R4B+LLaMA & 0.605 & 0.640 & 0.371 & 0.896 \\
Rerank (top-3) & valid & Stella+Qwen3-R4B+LLaMA & 0.618 & 0.643 & 0.393 & 0.899 \\
Rerank (top-3) & valid-1k & Stella+Qwen3-R4B+LLaMA & 0.621 & 0.640 & 0.373 & 0.895 \\
Rerank (top-3) & valid & Qwen3-E4B+Qwen3-R4B+Qwen3-30B & 0.600 & 0.753 & 0.567 & 0.944 \\
Rerank (top-3) & valid-1k & Qwen3-E4B+Qwen3-R4B+Qwen3-30B & 0.605 & 0.750 & 0.541 & 0.949 \\
Rerank (top-3) & valid & Stella+Qwen3-R4B+Qwen3-30B & 0.618 & 0.762 & 0.592 & 0.936 \\
Rerank (top-3) & valid-1k & Stella+Qwen3-R4B+Qwen3-30B & 0.621 & 0.748 & 0.564 & 0.924 \\
\midrule
Rerank (top-5) & valid & Qwen3-E4B+Qwen3-R4B+Qwen3-4B & 0.638 & 0.671 & 0.532 & 0.815 \\
Rerank (top-5) & valid-1k & Qwen3-E4B+Qwen3-R4B+Qwen3-4B & 0.643 & 0.650 & 0.492 & 0.801 \\
Rerank (top-5) & valid & Stella+Qwen3-R4B+Qwen3-4B & 0.648 & 0.674 & 0.543 & 0.808 \\
Rerank (top-5) & valid-1k & Stella+Qwen3-R4B+Qwen3-4B & 0.643 & 0.647 & 0.498 & 0.789 \\
Rerank (top-5) & valid & Qwen3-E4B+Qwen3-R4B+LLaMA & 0.638 & 0.647 & 0.409 & 0.892 \\
Rerank (top-5) & valid-1k & Qwen3-E4B+Qwen3-R4B+LLaMA & 0.643 & 0.643 & 0.385 & 0.889 \\
Rerank (top-5) & valid & Stella+Qwen3-R4B+LLaMA & 0.648 & 0.650 & 0.412 & 0.894 \\
Rerank (top-5) & valid-1k & Stella+Qwen3-R4B+LLaMA & 0.643 & 0.642 & 0.385 & 0.887 \\
Rerank (top-5) & valid & Qwen3-E4B+Qwen3-R4B+Qwen3-30B & 0.638 & 0.770 & 0.604 & 0.939 \\
Rerank (top-5) & valid-1k & Qwen3-E4B+Qwen3-R4B+Qwen3-30B & 0.643 & 0.760 & 0.561 & 0.949 \\
Rerank (top-5) & valid & Stella+Qwen3-R4B+Qwen3-30B & 0.648 & 0.774 & 0.616 & 0.935 \\
Rerank (top-5) & valid-1k & Stella+Qwen3-R4B+Qwen3-30B & 0.643 & 0.768 & 0.597 & 0.945 \\
\midrule
\ours{} (5-step) & valid-1k & Qwen3-4B & 0.528 & 0.630 & 0.459 & 0.793 \\
\ours{} (5-step) & valid-1k & LLaMA & 0.450 & 0.592 & 0.238 & 0.930 \\
\ours{} (5-step) & valid-1k & Qwen3-30B & 0.642 & 0.749 & 0.559 & 0.930 \\
\ours{} (10-step) & valid-1k & Qwen3-4B & 0.544 & 0.631 & 0.473 & 0.781 \\
\ours{} (10-step) & valid-1k & LLaMA & 0.482 & 0.593 & 0.256 & 0.914 \\
\ours{} (10-step) & valid-1k & Qwen3-30B & 0.671 & 0.762 & 0.590 & 0.926 \\
\bottomrule
\end{tabular}
\caption{Results on TARGET-TabFact-valid and TARGET-TabFact-valid-1k. Retrieval is R@1 for Dense/Rerank and R@all for \ours{}.}
\label{tab:main_tabfact_full2}
\end{table*}

As shown in Tables~\ref{tab:appendix_tcr_full} and~\ref{tab:appendix_tcr_others}, \ours{} exhibits substantial variation across backbone models. Unlike Dense and Rerank, whose retrieval quality is determined by fixed retrieval components, \ours{} relies on the backbone LLM to generate search patterns, interpret intermediate evidence, and decide when refinement is needed. As a result, retrieval effectiveness is closely tied to the backbone's ability to perform these search-related reasoning tasks. Across the evaluated models, higher retrieval performance generally corresponds to better end-to-end QA results. Notably, Qwen3.6-27B delivers the strongest overall performance, achieving both the highest retrieval score and the highest QA F1 among the evaluated backbones.

An interesting observation comes from the table-specialized model TableGPT-R1. Under the oracle setting, TableGPT-R1 outperforms several general-purpose models, indicating strong table understanding and reasoning capabilities once the correct table is provided. However, when used as the retrieval backbone of \ours{}, its retrieval performance remains below that of Qwen3.5-9B. This suggests that effective table reasoning and effective search planning may not be entirely the same capability. At the same time, given the limited number of table-specialized models currently available, it remains possible that future table-oriented foundation models could combine strong table understanding with stronger search and retrieval abilities.

The effect of search budget also varies across backbones. Some models obtain only limited gains when the budget increases from 5 to 10 steps, whereas others continue to benefit from additional refinement rounds. For example, GLM-Z1 improves from 0.258 to 0.306 in retrieval and from 0.170 to 0.180 in F1, while Qwen3.6-27B improves from 0.641 to 0.684 in retrieval and from 0.545 to 0.591 in F1. Qwen3.5-9B and TableGPT-R1 also benefit from additional refinement rounds, although to different extents. This indicates that the value of a larger search budget depends not only on the search procedure itself, but also on the backbone's ability to effectively utilize the additional opportunities for exploration and refinement.

Overall, these results suggest that \ours{} should be viewed as a retrieval framework whose effectiveness is tightly coupled to the capabilities of the underlying backbone model. When paired with a backbone that can effectively plan, refine, and interpret intermediate search results, \ours{} can substantially outperform conventional retrieval-based pipelines in both retrieval and downstream QA performance. Given that different backbones excel at different sub-tasks, a natural future direction is to explore model composition strategies, such as dynamic routing or fine-tuning specialized modules for each stage, to boost overall performance while maintaining deployment efficiency.

On \textsc{TARGET-TabFact-test-1k} (Table~\ref{tab:main_tabfact_full}; see Table~\ref{tab:main_tabfact_full2} for results on the valid set and its sampled subsets), \ours{} (10-step) attains the highest Acc (0.764) and Acc-Sup (0.589) among all methods, despite its retrieval score (0.648) being slightly below that of Rerank (top-5) (0.656). This mirrors the R@all-to-F1 conversion gap discussed above for TCR-Bench: a higher retrieval score does not necessarily translate into better downstream QA, and \ours{}'s tighter, refine-driven context appears to convert retrieval into correct answers more effectively than Rerank's fixed top-$k$ context. Acc-Ref is high and relatively similar across all methods, including Dense, consistent with the floor effect noted earlier: TabFact's ``not supported'' label is comparatively easy to predict even under weak or incorrect retrieval, so it is a less discriminative signal than Acc-Sup or overall Acc. Notably, even \ours{} (5-step) already matches Rerank (top-5) in Acc (0.754) while using a substantially smaller retrieval budget, suggesting that \ours{}'s advantage on TabFact is not solely a matter of achieving higher retrieval scores, but of making better use of what it retrieves.

\subsection{Statistical Significance Analysis}
\label{sec:appendix_significance}

To further assess the reliability of the observed improvements, we conduct paired significance tests on TCR-Bench using per-instance QA F1 scores. Following common practice in NLP and information retrieval, our primary test is paired bootstrap resampling with 10,000 samples. We additionally report Wilcoxon signed-rank and paired $t$-tests as supplementary references.

Table~\ref{tab:significance} compares \ours{} against the strongest reranking baselines under matched search budgets. We report results for both Qwen3-30B, which is used throughout the main experiments, and the stronger Qwen3.6-27B model. For Qwen3-30B, \ours{} consistently achieves higher mean F1 than the corresponding reranking baseline. Under the larger search budget, \ours{} (10-step) improves F1 from 0.317 to 0.361 ($\Delta=+0.045$), while under the smaller budget, \ours{} (5-step) improves F1 from 0.259 to 0.284 ($\Delta=+0.024$).

However, for Qwen3-30B, none of the statistical tests reject the null hypothesis at the 0.05 significance level. The 95\% bootstrap confidence intervals include zero in both comparisons, and the Wilcoxon and paired $t$-tests lead to the same conclusion. These results indicate that, while the observed improvements are consistently positive, the current evaluation size does not provide sufficient evidence to establish statistical superiority over the reranking baseline.

To examine whether stronger reasoning capability changes this conclusion, we additionally evaluate Qwen3.6-27B. Under the same retrieval settings, \ours{} substantially outperforms reranking baselines. The 10-step variant improves QA F1 from 0.359 to 0.592 ($\Delta=+0.233$), while the 5-step variant improves QA F1 from 0.323 to 0.545 ($\Delta=+0.222$). In both cases, the bootstrap confidence intervals are strictly positive and the corresponding $p$-values are below 0.001. We note that these gains come at a substantially higher computational cost: the full benchmark evaluation of the 10-step variant requires approximately 8.247 hours.

We additionally evaluate the effect of increasing the search budget within \ours{} using the Qwen3-30B backbone. Comparing the 5-step and 10-step variants, the mean QA F1 improves from 0.284 to 0.361 ($\Delta=+0.077$). In contrast to the reranking comparisons, all three significance tests reject the null hypothesis. The paired bootstrap test yields a 95\% confidence interval of [0.028, 0.128] and a two-sided $p$-value of 0.002. Wilcoxon signed-rank and paired $t$-tests produce consistent conclusions ($p<0.01$). These results provide evidence that increasing the search budget yields statistically reliable improvements within the proposed framework under the Qwen3-30B setting.

Overall, the significance analysis reveals three observations. First, under the Qwen3-30B setting used in the main paper, \ours{} consistently outperforms reranking baselines but does not achieve statistical significance at the conventional 0.05 level, suggesting performance comparable to strong retrieval-and-reranking pipelines rather than conclusive superiority. Second, increasing the search budget from 5-step to 10-step produces statistically significant gains for Qwen3-30B. Third, when paired with a substantially stronger reasoning model, \ours{} achieves large and statistically significant improvements over reranking baselines. Together, these findings suggest that the proposed framework can effectively benefit from stronger reasoning capabilities and may yield further gains as foundation models continue to improve.

\begin{table*}[t]
\centering
\small
\begin{tabular}{lcccccc}
\toprule
Comparison & Backbone LLM & F1$_A$ & F1$_B$ & $\Delta$ & 95\% CI & $p$ \\
\midrule
\ours{} (10-step) vs Rerank (top-5)
& Qwen3.6-27B
& 0.592
& 0.359
& +0.233
& [0.167, 0.302]
& $<0.001$ \\

\ours{} (5-step) vs Rerank (top-3)
& Qwen3.6-27B
& 0.545
& 0.323
& +0.222
& [0.152, 0.294]
& $<0.001$ \\

\ours{} (10-step) vs Rerank (top-5)
& Qwen3-30B
& 0.361
& 0.317
& +0.045
& [-0.028, 0.118]
& 0.229 \\

\ours{} (5-step) vs Rerank (top-3)
& Qwen3-30B
& 0.284
& 0.259
& +0.024
& [-0.044, 0.095]
& 0.482 \\

\midrule

\ours{} (10-step) vs \ours{} (5-step)
& Qwen3-30B
& 0.361
& 0.284
& +0.077
& [0.028, 0.128]
& 0.002 \\

\bottomrule
\end{tabular}
\caption{Paired significance analysis on TCR-Bench using per-instance QA F1 scores. Bootstrap uses 10,000 resamples.}
\label{tab:significance}
\end{table*}

\section{Oracle Results (Gold Table Only)}
\label{sec:Oracle_Results}

As an upper-bound reference, we first evaluate QA performance when the gold table is provided directly to the LLM, without any retrieval step.

\textbf{TCR-Bench.} Table~\ref{tab:oracle_tcr} reports oracle QA results under different table formats. Since retrieval is bypassed, these results reflect the models' intrinsic ability to understand and reason over table content when the correct evidence is already available.

\begin{table}[t]
\centering
\small
\begin{tabular}{lcccc}
\toprule
Model & CSV & Markdown & HTML & Mixed \\
\midrule
Qwen3-4B  & 0.239 & 0.301 & 0.266 & 0.270 \\
LLaMA     & 0.491 & 0.485 & 0.417 & 0.465 \\
Qwen3-30B & 0.717 & 0.748 & 0.719 & 0.734 \\
\bottomrule
\end{tabular}
\caption{Oracle QA results (F1) on TCR-Bench under different table formats.}
\label{tab:oracle_tcr}
\end{table}

\textbf{TabFact.} Table~\ref{tab:oracle_tabfact} reports results on TARGET-TabFact.

\begin{table}[t]
\centering
\begin{tabular}{llccc}
\toprule
Model & Split & Acc & Acc-Sup & Acc-Ref \\
\midrule
Qwen3-4B & valid & 0.732 & 0.690 & 0.775 \\
Qwen3-4B & valid-1k & 0.723 & 0.683 & 0.764 \\
Qwen3-4B & test & 0.695 & 0.654 & 0.737 \\
Qwen3-4B & test-1k & 0.695 & 0.654 & 0.737 \\
\midrule
LLaMA & valid & 0.699 & 0.528 & 0.875 \\
LLaMA & valid-1k & 0.684 & 0.488 & 0.871 \\
LLaMA & test & 0.696 & 0.513 & 0.881 \\
LLaMA & test-1k & 0.693 & 0.501 & 0.892 \\
\midrule
Qwen3-30B & valid & 0.904 & 0.855 & 0.953 \\
Qwen3-30B & valid-1k & 0.881 & 0.807 & 0.951 \\
Qwen3-30B & test & 0.907 & 0.859 & 0.955 \\
Qwen3-30B & test-1k & 0.896 & 0.835 & 0.959 \\
\bottomrule
\end{tabular}
\caption{Oracle QA results on TabFact.}
\label{tab:oracle_tabfact}
\end{table}

\subsection{No-Table QA}
\label{sec:no_table_qa}

To assess whether benchmark performance can be achieved from parametric knowledge alone, we evaluate a no-table setting on TCR-Bench, where the model receives only the question and no table content.

Performance remains close to zero across all formats, indicating that answering TCR-Bench questions depends primarily on access to table evidence rather than memorized knowledge. Specifically, on Qwen3-30B, we obtain F1 scores of 0.019 on CSV, 0.024 on Markdown, 0.019 on HTML, and 0.014 on Mixed format. We report Qwen3-30B only, as the purpose of this experiment is to provide a sanity check rather than a model comparison. We do not report analogous results on TabFact, since removing the table renders the task largely uninterpretable. With only two possible labels (true or false), model predictions are heavily confounded by dataset priors (e.g., a bias toward predicting false).


\section{Additional Retrieval-Only Results}
\label{app:retrieval_only}

The primary evaluation in this work focuses on end-to-end retrieval and question answering performance. Evaluating every retrieval paradigm under the complete pipeline would require additional QA inference and answer matching for each retriever configuration. Therefore, for completeness, we additionally report retrieval-only results for several representative retrieval methods on TCR-Bench.

Table~\ref{tab:retrieval_only}, Table~\ref{tab:retrieval_only_target_valid}, and Table~\ref{tab:retrieval_only_target_test} present retrieval recall under different retrieval settings. Following the standard retrieval evaluation protocol, a query is considered successfully retrieved if at least one evidence table appears within the top-$k$ retrieved results.

\begin{table*}[!tbp]
\centering
\small
\begin{tabular}{l l l c c c}
\toprule
Method & Model(s) & Format & R@1 & R@2 & R@3 \\
\midrule
Dense & Qwen3-Embedding-4B & CSV & 0.158 & 0.311 & 0.464 \\
Dense & Qwen3-Embedding-4B & Mixed & 0.172 & 0.321 & 0.469 \\
Dense & Qwen3-Embedding-4B & HTML & 0.134 & 0.287 & 0.474 \\
Dense & Qwen3-Embedding-4B & Markdown & 0.187 & 0.340 & 0.512 \\
\midrule
Dense & jina-embeddings-v4 & CSV & 0.129 & 0.282 & 0.368 \\
Dense & jina-embeddings-v4 & Mixed & 0.163 & 0.306 & 0.426 \\
Dense & jina-embeddings-v4 & HTML & 0.158 & 0.368 & 0.474 \\
Dense & jina-embeddings-v4 & Markdown & 0.139 & 0.268 & 0.373 \\
\midrule
Dense & stella\_en\_1.5B\_v5 & CSV & 0.129 & 0.244 & 0.335 \\
Dense & stella\_en\_1.5B\_v5 & Mixed & 0.096 & 0.172 & 0.244 \\
Dense & stella\_en\_1.5B\_v5 & HTML & 0.048 & 0.105 & 0.172 \\
Dense & stella\_en\_1.5B\_v5 & Markdown & 0.139 & 0.244 & 0.325 \\
\midrule
BM25 & BM25 & CSV & 0.010 & 0.010 & 0.014 \\
BM25 & BM25 & Mixed & 0.010 & 0.010 & 0.014 \\
BM25 & BM25 & HTML & 0.010 & 0.014 & 0.014 \\
BM25 & BM25 & Markdown & 0.010 & 0.010 & 0.014 \\
\midrule
SPLADE & splade-v3 & CSV & 0.077 & 0.158 & 0.211 \\
SPLADE & splade-v3 & Mixed & 0.091 & 0.172 & 0.244 \\
SPLADE & splade-v3 & HTML & 0.062 & 0.124 & 0.182 \\
SPLADE & splade-v3 & Markdown & 0.096 & 0.139 & 0.191 \\
\midrule
ColBERT & SauerkrautLM-Multi-ModernColBERT & CSV & 0.182 & 0.340 & 0.464 \\
ColBERT & SauerkrautLM-Multi-ModernColBERT & Mixed & 0.172 & 0.311 & 0.474 \\
ColBERT & SauerkrautLM-Multi-ModernColBERT & HTML & 0.172 & 0.340 & 0.483 \\
ColBERT & SauerkrautLM-Multi-ModernColBERT & Markdown & 0.187 & 0.359 & 0.459 \\
\midrule
Rerank (top-3) & Qwen3-Embedding-4B+Qwen3-Reranker-0.6B & CSV & 0.335 & 0.411 & 0.464 \\
Rerank (top-3) & Qwen3-Embedding-4B+Qwen3-Reranker-0.6B & Mixed & 0.325 & 0.421 & 0.469 \\
Rerank (top-3) & Qwen3-Embedding-4B+Qwen3-Reranker-0.6B & HTML & 0.368 & 0.426 & 0.474 \\
Rerank (top-3) & Qwen3-Embedding-4B+Qwen3-Reranker-0.6B & Markdown & 0.359 & 0.469 & 0.512 \\
\midrule
Rerank (top-3) & Qwen3-Embedding-4B+Qwen3-Reranker-4B & CSV & 0.354 & 0.431 & 0.464 \\
Rerank (top-3) & Qwen3-Embedding-4B+Qwen3-Reranker-4B & Mixed & 0.359 & 0.426 & 0.469 \\
Rerank (top-3) & Qwen3-Embedding-4B+Qwen3-Reranker-4B & HTML & 0.373 & 0.435 & 0.474 \\
Rerank (top-3) & Qwen3-Embedding-4B+Qwen3-Reranker-4B & Markdown & 0.407 & 0.474 & 0.512 \\
\midrule
Rerank (top-3) & Qwen3-Embedding-4B+jina-reranker-v3 & CSV & 0.244 & 0.359 & 0.464 \\
Rerank (top-3) & Qwen3-Embedding-4B+jina-reranker-v3 & Mixed & 0.234 & 0.354 & 0.469 \\
Rerank (top-3) & Qwen3-Embedding-4B+jina-reranker-v3 & HTML & 0.206 & 0.354 & 0.474 \\
Rerank (top-3) & Qwen3-Embedding-4B+jina-reranker-v3 & Markdown & 0.230 & 0.397 & 0.512 \\
\midrule
Rerank (top-5) & Qwen3-Embedding-4B+Qwen3-Reranker-0.6B & CSV & 0.416 & 0.526 & 0.603 \\
Rerank (top-5) & Qwen3-Embedding-4B+Qwen3-Reranker-0.6B & Mixed & 0.359 & 0.498 & 0.574 \\
Rerank (top-5) & Qwen3-Embedding-4B+Qwen3-Reranker-0.6B & HTML & 0.450 & 0.555 & 0.632 \\
Rerank (top-5) & Qwen3-Embedding-4B+Qwen3-Reranker-0.6B & Markdown & 0.402 & 0.522 & 0.589 \\
\midrule
Rerank (top-5) & Qwen3-Embedding-4B+Qwen3-Reranker-4B & CSV & 0.445 & 0.531 & 0.598 \\
Rerank (top-5) & Qwen3-Embedding-4B+Qwen3-Reranker-4B & Mixed & 0.421 & 0.526 & 0.584 \\
Rerank (top-5) & Qwen3-Embedding-4B+Qwen3-Reranker-4B & HTML & 0.459 & 0.560 & 0.608 \\
Rerank (top-5) & Qwen3-Embedding-4B+Qwen3-Reranker-4B & Markdown & 0.488 & 0.526 & 0.603 \\
\midrule
Rerank (top-5) & Qwen3-Embedding-4B+jina-reranker-v3 & CSV & 0.268 & 0.397 & 0.522 \\
Rerank (top-5) & Qwen3-Embedding-4B+jina-reranker-v3 & Mixed & 0.254 & 0.411 & 0.507 \\
Rerank (top-5) & Qwen3-Embedding-4B+jina-reranker-v3 & HTML & 0.234 & 0.388 & 0.512 \\
Rerank (top-5) & Qwen3-Embedding-4B+jina-reranker-v3 & Markdown & 0.230 & 0.411 & 0.541 \\
\bottomrule
\end{tabular}
\caption{Retrieval-only results on TCR-Bench.}
\label{tab:retrieval_only}
\end{table*}

\begin{table*}[!tbp]
\centering
\small
\begin{tabular}{l l l c c c}
\toprule
Method & Model(s) & Split & R@1 & R@2 & R@3 \\
\midrule
Dense & Qwen3-Embedding-4B & test & 0.495 & 0.596 & 0.649 \\
Dense & Qwen3-Embedding-4B & test-1k & 0.498 & 0.589 & 0.646 \\
Dense & jina-embeddings-v4 & test & 0.439 & 0.528 & 0.579 \\
Dense & jina-embeddings-v4 & test-1k & 0.450 & 0.542 & 0.593 \\
Dense & stella\_en\_1.5B\_v5 & test & 0.520 & 0.615 & 0.664 \\
Dense & stella\_en\_1.5B\_v5 & test-1k & 0.519 & 0.614 & 0.668 \\
\midrule
BM25 & BM25 & test & 0.104 & 0.147 & 0.165 \\
BM25 & BM25 & test-1k & 0.119 & 0.151 & 0.174 \\
\midrule
SPLADE & splade-v3 & test & 0.392 & 0.463 & 0.502 \\
SPLADE & splade-v3 & test-1k & 0.406 & 0.480 & 0.514 \\
\midrule
Rerank (top-3) & Qwen3-Embedding-4B+Qwen3-Reranker-0.6B & test & 0.593 & 0.634 & 0.649 \\
Rerank (top-3) & Qwen3-Embedding-4B+Qwen3-Reranker-0.6B & test-1k & 0.585 & 0.633 & 0.646 \\
Rerank (top-3) & stella\_en\_1.5B\_v5+Qwen3-Reranker-0.6B & test & 0.602 & 0.647 & 0.664 \\
Rerank (top-3) & stella\_en\_1.5B\_v5+Qwen3-Reranker-0.6B & test-1k & 0.606 & 0.653 & 0.668 \\
Rerank (top-3) & Qwen3-Embedding-4B+Qwen3-Reranker-4B & test & 0.603 & 0.637 & 0.649 \\
Rerank (top-3) & Qwen3-Embedding-4B+Qwen3-Reranker-4B & test-1k & 0.604 & 0.635 & 0.646 \\
Rerank (top-3) & stella\_en\_1.5B\_v5+Qwen3-Reranker-4B & test & 0.612 & 0.651 & 0.664 \\
Rerank (top-3) & stella\_en\_1.5B\_v5+Qwen3-Reranker-4B & test-1k & 0.623 & 0.654 & 0.668 \\
Rerank (top-3) & Qwen3-Embedding-4B+jina-reranker-v3 & test & 0.586 & 0.633 & 0.649 \\
Rerank (top-3) & Qwen3-Embedding-4B+jina-reranker-v3 & test-1k & 0.577 & 0.628 & 0.646 \\
Rerank (top-3) & stella\_en\_1.5B\_v5+jina-reranker-v3 & test & 0.592 & 0.645 & 0.664 \\
Rerank (top-3) & stella\_en\_1.5B\_v5+jina-reranker-v3 & test-1k & 0.592 & 0.649 & 0.668 \\
\midrule
Rerank (top-5) & Qwen3-Embedding-4B+Qwen3-Reranker-0.6B & test & 0.627 & 0.671 & 0.693 \\
Rerank (top-5) & Qwen3-Embedding-4B+Qwen3-Reranker-0.6B & test-1k & 0.615 & 0.660 & 0.684 \\
Rerank (top-5) & stella\_en\_1.5B\_v5+Qwen3-Reranker-0.6B & test & 0.627 & 0.678 & 0.701 \\
Rerank (top-5) & stella\_en\_1.5B\_v5+Qwen3-Reranker-0.6B & test-1k & 0.631 & 0.682 & 0.706 \\
Rerank (top-5) & Qwen3-Embedding-4B+Qwen3-Reranker-4B & test & 0.641 & 0.679 & 0.700 \\
Rerank (top-5) & Qwen3-Embedding-4B+Qwen3-Reranker-4B & test-1k & 0.635 & 0.669 & 0.688 \\
Rerank (top-5) & stella\_en\_1.5B\_v5+Qwen3-Reranker-4B & test & 0.647 & 0.690 & 0.710 \\
Rerank (top-5) & stella\_en\_1.5B\_v5+Qwen3-Reranker-4B & test-1k & 0.656 & 0.698 & 0.721 \\
Rerank (top-5) & Qwen3-Embedding-4B+jina-reranker-v3 & test & 0.614 & 0.671 & 0.695 \\
Rerank (top-5) & Qwen3-Embedding-4B+jina-reranker-v3 & test-1k & 0.604 & 0.659 & 0.686 \\
Rerank (top-5) & stella\_en\_1.5B\_v5+jina-reranker-v3 & test & 0.614 & 0.676 & 0.703 \\
Rerank (top-5) & stella\_en\_1.5B\_v5+jina-reranker-v3 & test-1k & 0.620 & 0.679 & 0.706 \\
\bottomrule
\end{tabular}
\caption{Retrieval-only results on TARGET-TabFact-test and TARGET-TabFact-test-1k.}
\label{tab:retrieval_only_target_test}
\end{table*}

\begin{table*}[!tbp]
\centering
\small
\begin{tabular}{l l l c c c}
\toprule
Method & Model(s) & Split & R@1 & R@2 & R@3 \\
\midrule
Dense & Qwen3-Embedding-4B & valid & 0.492 & 0.592 & 0.647 \\
Dense & Qwen3-Embedding-4B & valid-1k & 0.515 & 0.606 & 0.641 \\
Dense & jina-embeddings-v4 & valid & 0.428 & 0.519 & 0.570 \\
Dense & jina-embeddings-v4 & valid-1k & 0.432 & 0.522 & 0.579 \\
Dense & stella\_en\_1.5B\_v5 & valid & 0.516 & 0.618 & 0.668 \\
Dense & stella\_en\_1.5B\_v5 & valid-1k & 0.523 & 0.611 & 0.661 \\
\midrule
BM25 & BM25 & valid & 0.061 & 0.130 & 0.153 \\
BM25 & BM25 & valid-1k & 0.073 & 0.146 & 0.160 \\
\midrule
SPLADE & splade-v3 & valid & 0.405 & 0.478 & 0.519 \\
SPLADE & splade-v3 & valid-1k & 0.406 & 0.469 & 0.510 \\
\midrule
Rerank (top-3) & Qwen3-Embedding-4B+Qwen3-Reranker-0.6B & valid & 0.590 & 0.631 & 0.647 \\
Rerank (top-3) & Qwen3-Embedding-4B+Qwen3-Reranker-0.6B & valid-1k & 0.588 & 0.627 & 0.641 \\
Rerank (top-3) & stella\_en\_1.5B\_v5+Qwen3-Reranker-0.6B & valid & 0.606 & 0.652 & 0.668 \\
Rerank (top-3) & stella\_en\_1.5B\_v5+Qwen3-Reranker-0.6B & valid-1k & 0.606 & 0.647 & 0.661 \\
Rerank (top-3) & Qwen3-Embedding-4B+Qwen3-Reranker-4B & valid & 0.600 & 0.639 & 0.647 \\
Rerank (top-3) & Qwen3-Embedding-4B+Qwen3-Reranker-4B & valid-1k & 0.605 & 0.633 & 0.641 \\
Rerank (top-3) & stella\_en\_1.5B\_v5+Qwen3-Reranker-4B & valid & 0.618 & 0.659 & 0.668 \\
Rerank (top-3) & stella\_en\_1.5B\_v5+Qwen3-Reranker-4B & valid-1k & 0.621 & 0.649 & 0.661 \\
Rerank (top-3) & Qwen3-Embedding-4B+jina-reranker-v3 & valid & 0.579 & 0.633 & 0.647 \\
Rerank (top-3) & Qwen3-Embedding-4B+jina-reranker-v3 & valid-1k & 0.584 & 0.632 & 0.641 \\
Rerank (top-3) & stella\_en\_1.5B\_v5+jina-reranker-v3 & valid & 0.597 & 0.651 & 0.668 \\
Rerank (top-3) & stella\_en\_1.5B\_v5+jina-reranker-v3 & valid-1k & 0.596 & 0.644 & 0.661 \\
\midrule
Rerank (top-5) & Qwen3-Embedding-4B+Qwen3-Reranker-0.6B & valid & 0.621 & 0.670 & 0.689 \\
Rerank (top-5) & Qwen3-Embedding-4B+Qwen3-Reranker-0.6B & valid-1k & 0.606 & 0.659 & 0.675 \\
Rerank (top-5) & stella\_en\_1.5B\_v5+Qwen3-Reranker-0.6B & valid & 0.632 & 0.682 & 0.706 \\
Rerank (top-5) & stella\_en\_1.5B\_v5+Qwen3-Reranker-0.6B & valid-1k & 0.622 & 0.673 & 0.686 \\
Rerank (top-5) & Qwen3-Embedding-4B+Qwen3-Reranker-4B & valid & 0.638 & 0.683 & 0.695 \\
Rerank (top-5) & Qwen3-Embedding-4B+Qwen3-Reranker-4B & valid-1k & 0.633 & 0.663 & 0.675 \\
Rerank (top-5) & stella\_en\_1.5B\_v5+Qwen3-Reranker-4B & valid & 0.648 & 0.697 & 0.712 \\
Rerank (top-5) & stella\_en\_1.5B\_v5+Qwen3-Reranker-4B & valid-1k & 0.643 & 0.677 & 0.695 \\
Rerank (top-5) & Qwen3-Embedding-4B+jina-reranker-v3 & valid & 0.606 & 0.670 & 0.692 \\
Rerank (top-5) & Qwen3-Embedding-4B+jina-reranker-v3 & valid-1k & 0.598 & 0.655 & 0.676 \\
Rerank (top-5) & stella\_en\_1.5B\_v5+jina-reranker-v3 & valid & 0.617 & 0.680 & 0.704 \\
Rerank (top-5) & stella\_en\_1.5B\_v5+jina-reranker-v3 & valid-1k & 0.613 & 0.668 & 0.685 \\
\bottomrule
\end{tabular}
\caption{Retrieval-only results on TARGET-TabFact-valid and TARGET-TabFact-valid-1k.}
\label{tab:retrieval_only_target_valid}
\end{table*}

\section{Details of Discriminative Score Metrics}
\label{app:ds_details}

Following TCR-Bench~\citep{TCR-Bench}, we use Group Recall ($GR$) to distinguish topical relevance from exact evidence retrieval. Each query is associated with a Sibling Table Group, which contains the gold table together with semantically related tables covering similar topics. A retrieval is counted as a group hit if it retrieves any table from this group, regardless of whether the retrieved table is the gold table itself.

For single-pass retrieval methods, we compute
\[
GR@1,
\]
which measures whether the retrieved top-ranked table belongs to the query's Sibling Table Group. The corresponding Discriminative Score is

\[
DS@1 = \frac{R@1}{GR@1}.
\]

Intuitively, $GR@1$ measures whether retrieval reaches the correct topical neighborhood, while $DS@1$ measures how often the method successfully identifies the gold table after reaching that neighborhood.

Extending this evaluation to \ours{} is less straightforward because retrieval occurs over multiple QA calls rather than a single ranked retrieval step. We therefore define two variants.

\paragraph{Retrieval-wide metrics}
Let $R@all$ denote whether the gold table is retrieved at least once across all QA calls within a trajectory. We define $GR@all$ analogously. If the gold table is retrieved anywhere in the trajectory, the query trivially counts as a group hit. Otherwise, we examine the final table used for answer generation and count it as a group hit if it belongs to the query's Sibling Table Group. The corresponding discriminative score is

\[
DS@all = \frac{R@all}{GR@all}.
\]

This metric evaluates whether the overall search process is capable of locating the correct evidence while accounting for semantically related alternatives encountered during the trajectory.

\paragraph{Final-decision metrics}
Since \ours{} ultimately produces an answer based on a single table, we additionally evaluate only the final retrieval decision. Let $R@final$ denote whether the final table used for answer generation is the gold table. Likewise, $GR@final$ measures whether that final table belongs to the Sibling Table Group. We then compute

\[
DS@final = \frac{R@final}{GR@final}.
\]

Unlike $R@all$, $R@final$ does not give credit for intermediate retrieval successes that are later discarded. It therefore reflects the quality of the final evidence selection decision made by the agent.

\paragraph{Interpretation}
Recall and Group Recall capture different aspects of retrieval quality. High $GR$ indicates that retrieval reaches the correct semantic neighborhood, whereas high $R$ requires identifying the exact answer-bearing table. Consequently, a high Discriminative Score indicates that retrieval errors are less frequently caused by semantically similar sibling tables and that the method is more effective at prioritizing evidence sufficiency over topical relevance alone.

\section{Full Efficiency Breakdown: Calls and Runtime}
\label{app:full_efficiency}
Table~\ref{tab:calls_time_full} reports the complete breakdown of calls and wall-clock runtime (hours) for each method on TCR-Bench, using Qwen3-4B and Qwen3-30B as the QA backbone where applicable.

\begin{table*}[t]
\centering
\setlength{\tabcolsep}{1mm} 
\small
\begin{tabular}{lcccccccc}
\toprule
Method & Dense Calls & Rerank Calls & LLM Calls & Total Calls & Dense Time & Rerank Time & LLM Time & Total Time \\
\midrule
Dense (Qwen3-4B)    & 846 & 0    & 209 & 1055 & 0.940 & 0     & 0.125 & 1.065 \\
Rerank-3 (Qwen3-4B) & 846 & 627  & 209 & 1682 & 0.940 & 0.964 & 0.119 & 2.023 \\
Rerank-5 (Qwen3-4B) & 846 & 1045 & 209 & 2100 & 0.940 & 1.597 & 0.122 & 2.659 \\
Dense (Qwen3-30B)    & 846 & 0    & 209 & 1055 & 0.940 & 0     & 0.765 & 1.705 \\
Rerank-3 (Qwen3-30B) & 846 & 627  & 209 & 1682 & 0.940 & 0.964 & 0.711 & 2.615 \\
Rerank-5 (Qwen3-30B) & 846 & 1045 & 209 & 2100 & 0.940 & 1.597 & 0.704 & 3.241 \\
\midrule
5-step (Qwen3-4B)    & 0 & 0 & 918  & 918  & 0 & 0 & 0.284 & 0.284 \\
10-step (Qwen3-4B)   & 0 & 0 & 1581 & 1581 & 0 & 0 & 0.599 & 0.599 \\
5-step (Qwen3-30B)   & 0 & 0 & 921  & 921  & 0 & 0 & 2.335 & 2.335 \\
10-step (Qwen3-30B)  & 0 & 0 & 1305 & 1305 & 0 & 0 & 3.466 & 3.466 \\
\bottomrule
\end{tabular}
\caption{Number of inference calls and wall-clock runtime (hours) on TCR-Bench.}
\label{tab:calls_time_full}
\end{table*}

\section{Fine-Grained Ablation of the Refine Mechanism}
\label{app:refine_ablation}

The refine mechanism in \ours{} operates through two sub-steps at each iteration: \texttt{regenerate}, which produces an updated set of search patterns and launches a new retrieval round to obtain a fresh candidate pool, and \texttt{select}, which decides whether the current candidate pool is already sufficient and can be passed on for verification. To better understand how each sub-step contributes to overall performance, we conduct a fine-grained ablation that isolates the two components separately, in addition to the full removal of refine studied in the main paper.

We consider three variants. In \texttt{w/o refine}, the refine mechanism is disabled entirely, so the agent proceeds directly from the initial retrieval round to answerability checking. In \texttt{w/o regenerate}, the agent is confined to the candidate pool obtained from the initial retrieval round and can only repeatedly apply select over this fixed pool, never generating new search patterns or triggering another retrieval round. In \texttt{w/o select}, the select step is removed, so the agent always regenerates new search patterns and launches another retrieval round, regardless of whether the current pool would already have been judged sufficient. This variant only stops for two reasons: the step budget is exhausted, at which point QA is performed directly on the current top-1 retrieved table, or \texttt{refine-none} occurs. All variants are evaluated on TCR-Bench (Mixed format) using Qwen3-30B under both 5-step and 10-step budgets, following the same experimental setup as the main ablation.

\begin{table}[t]
\centering
\small
\begin{tabular}{lcccc}
\toprule
Strategy & R@all & F1 & Avg.\ Steps & Time (h) \\
\midrule
\textbf{5-step setting} & & & & \\
\midrule
Full & \textbf{0.416} & 0.284 & 4.407 & 2.335 \\
w/o refine & 0.239 & 0.173 & 3.555 & 1.733 \\
w/o regenerate & 0.254 & 0.190 & 4.148 & 1.921 \\
w/o select & 0.402 & \textbf{0.305} & 4.909 & 3.031 \\
\midrule
\textbf{10-step setting} & & & & \\
\midrule
Full & 0.507 & 0.361 & 6.244 & 3.466 \\
w/o refine & 0.268 & 0.195 & 5.019 & 2.408 \\
w/o regenerate & 0.278 & 0.202 & 5.823 & 2.712 \\
w/o select & \textbf{0.536} & \textbf{0.398} & 8.167 & 5.638 \\
\bottomrule
\end{tabular}
\caption{Fine-grained ablation of the refine mechanism, decomposed into regenerate and select, on TCR-Bench (Mixed format) using Qwen3-30B.}
\label{tab:refine_finegrounded_ablation}
\end{table}

Table~\ref{tab:refine_finegrounded_ablation} shows that the two sub-steps contribute in clearly different ways, and are not interchangeable. Removing regenerate causes a consistent drop in both R@all and F1 relative to the full model under both budgets, since the agent can never move beyond the initial retrieval round and is limited to selecting among a fixed candidate pool that may not cover the evidence needed to answer the query. Nonetheless, \texttt{w/o regenerate} still clearly outperforms \texttt{w/o refine} under both budgets (e.g.\ R@all/F1 of 0.254/0.190 versus 0.239/0.173 under 5 steps, and 0.278/0.202 versus 0.268/0.195 under 10 steps), indicating that even without the ability to launch new retrieval rounds, the select step alone still provides some benefit over having no refinement at all by allowing the agent to make a more deliberate choice among the initially retrieved candidates.

Removing select, in contrast, does not lead to a comparable degradation: F1 actually improves over the full model in both settings (0.305 vs.\ 0.284 under 5 steps, 0.398 vs.\ 0.361 under 10 steps), and R@all is also higher under 10 steps. The gain, however, comes at a substantially higher cost. Under 10 steps, the average number of steps for \texttt{w/o select} rises to 8.167, still below the nominal budget of 10 since a fraction of runs terminate early through \texttt{refine-none} rather than running the full budget to a forced QA, but this remains far above the 6.244 steps used by the full model, and the corresponding runtime increases by a factor of 1.63 (from 3.466 to 5.638 hours).

These results indicate that regenerate is the component primarily responsible for refine's improvement over \texttt{w/o refine}, while select trades a modest amount of F1 for a substantial reduction in average steps and runtime by allowing the agent to stop generating new search patterns once the current candidate pool is already sufficient. Since \ours{} is designed to balance retrieval quality with inference efficiency rather than pursue accuracy alone, we retain select in the full model as a favorable trade-off between the two objectives.

\section{Sustained Improvement Across Rounds}
\label{app:Sustained_Improvement}

The full results in Table~\ref{tab:refine-rounds-full} confirm that refinement is not a one-shot correction: for Qwen3-30B the hit rate improves in nearly every round from round~1 through round~7 (0.301, 0.330, 0.340, 0.364, 0.383, 0.378, 0.392), with only a single minor dip at round~6, before flattening between rounds~7 and~8. This near-monotonic trend across seven refinement steps indicates that each additional round provides new, usable signal rather than the model simply reproducing its earlier candidates. Qwen3.5-9B exhibits the same qualitative trend but reaches saturation much earlier, flattening after round~5.

\subsection{Capacity-Dependent Magnitude}

The extent of this sustained gain is capability-dependent. Qwen3-4B reaches its ceiling essentially at round~1 (0.158, versus 0.153 at round~0) and remains nearly unchanged thereafter, staying within 0.005 of that value through round~8. Qwen3.5-9B exhibits a modestly stronger refinement effect: its hit rate rises from 0.306 at round~0 to 0.349 at round~1 and reaches 0.368 by round~5, after which performance plateaus. While this demonstrates that iterative refinement remains beneficial at a medium model scale, the improvement is still substantially smaller than that observed for Qwen3-30B. The latter continues improving through most refinement rounds, gaining from 0.201 at round~0 to 0.392 at round~7. The TabFact subset shows an intermediate pattern: most of the gain is realized by round~2, with only small further improvements afterward, which is consistent with the possibility that TabFact's simpler table schemas offer less scope for iterative refinement.

GLM-Z1 shows a distinct profile from all three. Its round-0 hit rate is by far the lowest of any configuration (0.062), but the single largest jump in the entire table occurs between round~0 and round~1, where GLM-Z1 more than triples its hit rate (0.062 to 0.206). After this initial jump, further gains are small and gradual: the hit rate edges from 0.206 at round~1 to 0.225 at round~8, with brief plateaus (rounds~1--2) and a minor dip at rounds~5--6 before recovering. This is the most front-loaded pattern observed, even more so than TabFact: essentially all of the practically meaningful gain is captured in the first refinement step, and rounds~2 through~8 contribute a comparatively small additional 0.019 on top of it. This suggests that for GLM-Z1, the first refinement round corrects a large share of recoverable errors, while later rounds mainly make marginal, incremental corrections.

\subsection{Run-to-Run Variance}

Because the 5-step and 10-step runs are independent executions rather than a single run truncated at different lengths, their overlapping rounds (0 to 3) let us estimate run-to-run variance directly. For Qwen3-4B, the two runs agree on every overlapping round, differing by at most 0.005 and matching exactly at rounds~0, 1, and~2. For Qwen3-30B, the same rounds show larger discrepancies (e.g., 0.311 versus 0.330 at round~2, a gap of 0.019), and the TabFact runs show a comparable spread. We attribute this to the longer chain-of-thought reasoning Qwen3-30B produces during pattern generation and refinement, which introduces more sampling variance across independent runs than the shorter, more deterministic outputs of Qwen3-4B. This indicates that reported hit rates for larger reasoning models should be read with this run-to-run spread in mind, rather than treated as exact values.

\begin{table*}[ht]
\centering
\small
\begin{tabular}{c cc cc cc c c}
\toprule
\multirow{2}{*}{Round} & \multicolumn{2}{c}{Qwen3-30B} & \multicolumn{2}{c}{Qwen3-4B} & \multicolumn{2}{c}{Qwen3-30B (TabFact)} & GLM-Z1 & Qwen3.5-9B \\
\cmidrule(lr){2-3} \cmidrule(lr){4-5} \cmidrule(lr){6-7} \cmidrule(lr){8-8} \cmidrule(lr){9-9}
 & 5-step & 10-step & 5-step & 10-step & 5-step & 10-step & 10-step & 10-step \\
\midrule
0 & 0.206 & 0.201 & 0.153 & 0.153 & 0.441 & 0.444 & 0.062 & 0.306 \\
1 & 0.301 & 0.301 & 0.158 & 0.158 & 0.488 & 0.486 & 0.206 & 0.349 \\
2 & 0.311 & 0.330 & 0.158 & 0.158 & 0.498 & 0.493 & 0.206 & 0.359 \\
3 & 0.330 & 0.340 & 0.163 & 0.163 & 0.505 & 0.493 & 0.210 & 0.359 \\
4 & --    & 0.364 & --    & 0.158 & --    & 0.506 & 0.220 & 0.364 \\
5 & --    & 0.383 & --    & 0.158 & --    & 0.501 & 0.215 & 0.368 \\
6 & --    & 0.378 & --    & 0.158 & --    & 0.503 & 0.215 & 0.368 \\
7 & --    & 0.392 & --    & 0.158 & --    & 0.505 & 0.220 & 0.368 \\
8 & --    & 0.392 & --    & 0.158 & --    & 0.507 & 0.225 & 0.368 \\
\bottomrule
\end{tabular}
\caption{Top-1 hit rate by refinement round, for two independent runs
(5-step and 10-step budgets) of each model configuration.}
\label{tab:refine-rounds-full}
\end{table*}

\section{Refine Strategies}
\label{app:Refine_Strategies}
Here we study the effect of the refine strategy itself, with the row cap fixed at \textit{small} throughout. We consider two refine strategies: the select-based $n$s$m$ strategy (select $m$ out of $n$ candidate tables), adopted as default in the main experiments, and a rerank-based alternative, denoted r$n$, in which the LLM may either regenerate grep patterns or directly rerank the existing $n$ retrieved tables during refinement.

All experiments in this section use Qwen3-30B on TCR-Bench (Mixed format). Table~\ref{tab:refine_ablation} reports R@all, F1, the R@all-to-F1 conversion efficiency (Eff $=$ F1/R@all), the average number of steps used, and wall-clock runtime for the $n$s$m$ strategy.

\begin{table*}[t]
\centering
\small
\begin{tabular}{lccccc}
\toprule
Strategy & R@all & F1 & Eff & Avg.\ Steps & Time (h) \\
\midrule
5step-5s1 & 0.416 & 0.284 & 0.683 & 4.407 & 2.335 \\
5step-5s3 & 0.445 & 0.307 & 0.690 & 4.407 & 2.609 \\
5step-10s1 & 0.397 & 0.294 & \textbf{0.741} & 4.383 & 2.807 \\
5step-10s3 & 0.431 & 0.297 & 0.689 & 4.407 & 3.129 \\
5step-10s5 & \textbf{0.469} & \textbf{0.316} & 0.674 & 4.354 & 3.259 \\
\midrule
10step-5s1 & 0.507 & \textbf{0.361} & \textbf{0.712} & 6.244 & 3.466 \\
10step-5s3 & \textbf{0.550} & 0.352 & 0.640 & 6.139 & 3.703 \\
10step-10s1 & 0.469 & 0.332 & 0.708 & 6.072 & 3.946 \\
10step-10s3 & 0.512 & 0.340 & 0.664 & 6.024 & 4.282 \\
10step-10s5 & 0.541 & 0.358 & 0.662 & 5.952 & 4.424 \\
\bottomrule
\end{tabular}
\caption{Effect of refine strategy on TCR-Bench (Mixed format), using Qwen3-30B. Eff $=$ F1/R@all.}
\label{tab:refine_ablation}
\end{table*}

At the 5-step budget, increasing either the candidate pool size $n$ or the number of retained tables $m$ generally improves retrieval coverage. For example, within the 10s$m$ family, R@all increases from 0.397 (10s1) to 0.469 (10s5), while F1 also rises from 0.294 to 0.316. However, the gains are not monotonic across all settings. Enlarging the candidate pool may introduce more semantically related but non-evidential tables, making the final selection more difficult. As a result, 10s1 yields lower R@all than 5s1 despite examining more candidates.

A similar trade-off appears at the 10-step budget. Retaining more tables consistently improves retrieval coverage, with R@all increasing from 0.507 (5s1) to 0.550 (5s3) and from 0.469 (10s1) to 0.541 (10s5). Nevertheless, higher retrieval coverage does not always translate into better QA performance. While 10step-5s3 achieves the highest R@all, its F1 remains slightly below that of 10step-5s1. This observation is consistent with prior findings that providing more retrieved context can introduce distracting information and reduce answer generation quality~\citep{RAG-more-context1,RAG-more-context2-Re-rag,RAG-more-context3-lost-in-middle,RAG-more-context4}.

The conversion efficiency metric further highlights this effect. In both the 5-step and 10-step settings, strategies that retain fewer tables generally achieve higher Eff. For instance, 5step-10s1 attains the highest efficiency (0.741) among all 5-step configurations, while 10step-5s1 achieves the highest efficiency (0.712) among the 10-step configurations. In contrast, larger-$m$ variants obtain higher R@all but convert retrieval gains into QA improvements less effectively.

Runtime increases steadily with both $n$ and $m$, whereas the average number of interaction steps remains relatively stable. This suggests that the additional cost mainly comes from processing larger candidate sets and longer QA contexts rather than from substantially deeper interaction trajectories.

Overall, larger candidate pools and larger retained sets tend to improve retrieval coverage, but the resulting QA gains are considerably smaller. Since higher R@all does not necessarily yield higher F1 and incurs additional computational cost, we adopt 5s1 as the default refine strategy in the main experiments and additionally report representative larger-$m$ settings to illustrate the retrieval-generation trade-off.

Table~\ref{tab:refine_rerank} reports the same set of metrics for the rerank-based r$n$ strategy, under the same setup as above.

\begin{table}[t]
\centering
\setlength{\tabcolsep}{1.5mm} 
\small
\begin{tabular}{lccccc}
\toprule
Strategy & R@all & F1 & Eff & Avg.\ Steps & Time (h) \\
\midrule
5step-r3  & 0.378 & 0.269 & 0.712 & 4.431 & 2.172 \\
5step-r5  & \textbf{0.416} & 0.291 & 0.700 & 4.459 & 2.580 \\
5step-r10 & \textbf{0.416} & \textbf{0.307} & \textbf{0.738} & 4.349 & 2.850 \\
\midrule
10step-r3 & 0.498 & 0.375 & 0.753 & 6.196 & 3.174 \\
10step-r5 & 0.550 & 0.384 & 0.698 & 6.067 & 3.535 \\
10step-r10 & 0.517 & 0.374 & 0.723 & 6.038 & 4.069 \\
\bottomrule
\end{tabular}
\caption{Effect of the rerank-based refine strategy (r$n$) on TCR-Bench (Mixed format), using Qwen3-30B. Eff $=$ F1/R@all.}
\label{tab:refine_rerank}
\end{table}

The r$n$ strategy exhibits a similar retrieval-coverage trade-off to $n$s$m$, although the trend is less monotonic. At the 5-step budget, increasing $n$ generally improves F1, with 5step-r10 achieving the best overall balance among the rerank variants. At the 10-step budget, 10step-r5 attains the highest R@all (0.550) and F1 (0.384), while both smaller and larger candidate pools perform slightly worse. This suggests that reranking benefits from a sufficiently diverse candidate set, but excessively large candidate pools may introduce additional distractors that reduce selection quality. Notably, all 10-step rerank variants achieve higher Eff than their corresponding high-recall select-based counterparts, indicating that reranking can convert retrieval gains into downstream QA improvements more effectively when retrieval coverage is already adequate.

Comparing the two strategies overall, neither consistently dominates the other across formats or metrics; Table~\ref{tab:refine_format} breaks down selected 10-step configurations by table format.

\begin{table}[t]
\centering
\begin{tabular}{lccc}
\toprule
Format & 10step-5s1 & 10step-5s3 & 10step-r5 \\
\midrule
CSV & 0.346 & 0.367 & 0.362 \\
Mixed & 0.361 & 0.352 & 0.384 \\
HTML & 0.364 & 0.356 & 0.352 \\
Markdown & 0.364 & 0.389 & 0.346 \\
\bottomrule
\end{tabular}
\caption{F1 of selected 10-step refine strategies across table formats on TCR-Bench, using Qwen3-30B.}
\label{tab:refine_format}
\end{table}

We adopt the select-based strategy in our main experiments primarily for its greater configuration flexibility: it decouples the number of candidates considered ($n$) from the number retained ($m$), whereas r$n$ conflates the two. Although rerank-based refinement can achieve stronger performance under larger step budgets, the select-based formulation enables a broader exploration of retrieval-generation trade-offs. Overall, we select the 5s1 configuration as our default, as it offers a favorable balance of runtime, retrieval effectiveness, and downstream QA performance.

\section{Detailed Results for Table Format Robustness}
\label{app:Detailed_Robustness}

\subsection{Quantitative Results}

Table~\ref{tab:robustness} provides the complete numerical results for the table format robustness experiment on TCR-Bench, supplementing the visualization in the main paper. For each method, we report the mean F1 across four formats, along with the range, standard deviation, and Format Robustness Score (FRS).

We first define the worst-case degradation (WCD) as the largest relative performance drop caused by an unfavorable table format:

\[
\mathrm{WCD} = \frac{\max(\mathrm{F1}) - \min(\mathrm{F1})}{\max(\mathrm{F1})}
\]

We then define the Format Robustness Score (FRS) as

\[
\mathrm{FRS}
=
100 \times (1-\mathrm{WCD})
=
100 \times
\frac{\min(\mathrm{F1})}
{\max(\mathrm{F1})}.
\]

A higher FRS indicates stronger robustness to table serialization changes, with 100\% corresponding to complete format invariance.

\begin{table}[t]
\centering
\setlength{\tabcolsep}{1mm} 
\small
\begin{tabular}{lccccc}
\toprule
Method & 
Dense & 
\makecell{Rerank\\(top-3)} & 
\makecell{Rerank\\(top-5)} & 
\makecell{\ours{}\\(5step)} & 
\makecell{\ours{}\\(10step)} \\
\midrule
CSV & 0.116 & 0.284 & 0.356 & 0.283 & 0.346 \\
Mixed & 0.118 & 0.269 & 0.317 & 0.284 & 0.361 \\
HTML & 0.122 & 0.279 & 0.353 & 0.292 & 0.364 \\
Markdown & 0.134 & 0.314 & 0.362 & 0.318 & 0.364 \\
\midrule
Mean & 0.123 & 0.287 & 0.347 & 0.294 & 0.359 \\
Range & 0.018 & 0.045 & 0.045 & 0.035 & 0.018 \\
Std.\ Dev.\ & 0.0070 & 0.0168 & 0.0176 & 0.0141 & 0.0075 \\
FRS & 86.6\% & 85.7\% & 87.6\% & 89.0\% & 95.1\% \\
\bottomrule
\end{tabular}
\caption{F1 across table formats and robustness statistics on TCR-Bench, using Qwen3-30B.}
\label{tab:robustness}
\end{table}

\subsection{Multi-Run Robustness Evaluation}

To assess the stability of the robustness analysis, we independently repeat the experiments three times for Rerank (top-5) and \ours{} (10-step). We first average the F1 scores across runs for each table format and then compute robustness statistics from the averaged results.

\begin{table*}[t]
\centering
\setlength{\tabcolsep}{3pt}
\begin{tabular}{l|cccc|cccc}
\toprule
&
\multicolumn{4}{c|}{\ours{} (10-step)} &
\multicolumn{4}{c}{Rerank (top-5)} \\
\cmidrule(lr){2-5}
\cmidrule(lr){6-9}
Format &
Run1 & Run2 & Run3 & Avg &
Run1 & Run2 & Run3 & Avg \\
\midrule
CSV & 0.346 & 0.340 & 0.340 & 0.342 & 0.356 & 0.351 & 0.352 & 0.353 \\
Mixed & 0.361 & 0.360 & 0.363 & 0.362 & 0.317 & 0.312 & 0.317 & 0.315 \\
HTML & 0.364 & 0.364 & 0.364 & 0.364 & 0.353 & 0.353 & 0.353 & 0.353 \\
Markdown & 0.364 & 0.358 & 0.375 & 0.366 & 0.362 & 0.362 & 0.362 & 0.362 \\
\midrule
Mean & -- & -- & -- & 0.358 & -- & -- & -- & 0.346 \\
Range & -- & -- & -- & 0.024 & -- & -- & -- & 0.047 \\
Std.\ Dev.\ & -- & -- & -- & 0.0095 & -- & -- & -- & 0.0180 \\
FRS & -- & -- & -- & 93.5\% & -- & -- & -- & 87.1\% \\
\bottomrule
\end{tabular}
\caption{Multi-run robustness analysis for Rerank (top-5) and \ours{} (10-step). Robustness statistics are computed from the format-wise average F1 scores across three independent runs.}
\label{tab:robustness_multirun}
\end{table*}

The results remain consistent with the main findings. After averaging across multiple runs, \ours{} (10-step) still achieves a substantially higher robustness score than Rerank (top-5) (93.5\% vs. 87.1\%). In addition, Rerank (top-5) exhibits a noticeably larger performance drop on the Mixed-format corpus, whereas the performance of \ours{} remains relatively stable across all four serialization formats. These results indicate that the robustness advantage of \ours{} remains stable across repeated executions.

\subsection{Robustness of Alternative Refine Strategies}

We further analyze the robustness of several alternative refine strategies introduced in the ablation study (Table~\ref{tab:refine_ablation}). The goal of this experiment is to examine whether modifications to the refine behavior substantially affect robustness under different table serialization formats.

\begin{table}[t]
\centering
\small
\begin{tabular}{lcc}
\toprule
Method & 10-step-5s3 & 10-step-r5 \\
\midrule
CSV & 0.367 & 0.362 \\
Mixed & 0.352 & 0.384 \\
HTML & 0.356 & 0.352 \\
Markdown & 0.389 & 0.346 \\
\midrule
Mean & 0.366 & 0.361 \\
Range & 0.037 & 0.038 \\
Std.\ Dev.\ & 0.0144 & 0.0145 \\
FRS & 90.5\% & 90.1\% \\
\bottomrule
\end{tabular}
\caption{Robustness analysis of alternative refine strategies.}
\label{tab:refine_robustness}
\end{table}

As shown in Table~\ref{tab:refine_robustness}, both alternative refine strategies achieve robustness scores above 90\%, with highly similar ranges and standard deviations. The differences between configurations are relatively small compared with the overall robustness gains observed over conventional retrieval baselines. Within the limited range of refine strategy variations examined under the 10-step setting, moderate adjustments exhibit limited impact on format robustness, suggesting that this robustness is a stable property of the overall iterative retrieval-and-refinement framework rather than a consequence of any specific variant.

\section{Effect of Retrieved Evidence Size}
\label{sec:appendix_size}

In \ours{}, table inspection is performed through lightweight row-level retrieval rather than loading entire tables by default. To examine the trade-off between evidence completeness and efficiency, we vary the amount of table content exposed to the LLM during retrieval.

Table~\ref{tab:size_analysis} compares four evidence-size configurations. The \textit{small}, \textit{medium}, and \textit{big} settings differ in the number of matched rows returned by the grep module, while the \textit{full} setting directly provides the entire table and does not use row matching.

Specifically, \textit{small} returns at most one row per pattern and at most five matched rows in total; \textit{medium} returns up to two rows per pattern and ten rows in total; \textit{big} returns up to three rows per pattern and fifteen rows in total. The \textit{full} configuration exposes the complete table content to the model.

Table~\ref{tab:size_analysis} reports results on the Mixed format of TCR-Bench using Qwen3-30B. Increasing the amount of retrieved evidence generally improves retrieval performance. The \textit{full} configuration achieves the highest retrieval scores under both 5-step and 10-step budgets. However, these gains come at a substantial computational cost. Compared with \textit{big}, the \textit{full} setting requires approximately three times longer runtime while providing only modest improvements in retrieval and QA performance.

Among the row-retrieval configurations, \textit{big} achieves the strongest retrieval performance, while requiring only a small fraction of the runtime of \textit{full}. The default \textit{small} setting sacrifices some retrieval accuracy but offers a reasonable efficiency-performance trade-off and is therefore used throughout the main experiments.

\begin{table}[t]
\centering
\small
\begin{tabular}{lcccc}
\toprule
Configuration & R@All & F1 & Avg. Steps & Time (h) \\
\midrule
\textbf{10-step setting} & & & & \\
\midrule
full   & 0.512 & 0.371 & 6.153 & 10.079 \\
big    & 0.502 & 0.370 & 6.139 & 3.498 \\
medium & 0.493 & 0.372 & 6.340 & 3.606 \\
small  & 0.507 & 0.361 & 6.244 & 3.466 \\
\midrule
\textbf{5-step setting} & & & & \\
\midrule
full   & 0.455 & 0.333 & 4.311 & 7.094 \\
big    & 0.431 & 0.319 & 4.397 & 2.406 \\
medium & 0.426 & 0.306 & 4.435 & 2.343 \\
small  & 0.416 & 0.284 & 4.407 & 2.335 \\
\bottomrule
\end{tabular}
\caption{Effect of retrieved evidence size on TCR-Bench (Mixed format) using Qwen3-30B. Larger evidence windows generally improve retrieval performance but incur substantially higher runtime costs.}
\label{tab:size_analysis}
\end{table}

\section{Transposed Tables: A Stress Test Beyond Row-Oriented Entity Structure}
\label{app:transposed_tables}

To further stress-test retrieval under more challenging table organizations, we evaluate all methods on a transposed version of TCR-Bench, where every table is transposed before indexing and retrieval. This setting intentionally weakens the common assumption that entities are organized by rows, since entity attributes are now distributed across columns. As a result, many row-oriented retrieval heuristics become less effective, making retrieval substantially more difficult.

It is worth noting that transposition also increases the effective input length for many tables because original row identifiers often become column headers after transposition. Consequently, reranking inputs become significantly longer. In preliminary experiments, the original 4B reranker frequently encountered out-of-memory issues in this setting. We therefore replace it with the smaller R0.6B reranker. Since this model has slightly lower retrieval performance than the default reranker, the results in this section should be viewed primarily as a reference stress test rather than a directly comparable benchmark.

\begin{table*}[t]
\centering
\begin{tabular}{llcc}
\toprule
Method & Config & Recall$_T$ & F1$_T$ \\
\midrule
Oracle & Qwen3-30B & 1.000 & 0.269 \\
\midrule
Dense & E4B+Qwen3-30B & 0.081 & 0.048 \\
Rerank (top-3) & E4B+R0.6B+Qwen3-30B & 0.187 & 0.076 \\
Rerank (top-5) & E4B+R0.6B+Qwen3-30B & 0.239 & 0.100 \\
\midrule
\ours{} (5-step-5s1-small) & Qwen3-30B & 0.263 & 0.110 \\
\ours{} (10-step-5s1-small) & Qwen3-30B & \textbf{0.335} & \textbf{0.139} \\
\bottomrule
\end{tabular}
\caption{Results on the transposed-table version of TCR-Bench.}
\label{tab:transposed_results}
\end{table*}

Table~\ref{tab:transposed_results} shows that retrieval becomes substantially harder after transposition. Dense retrieval suffers the largest degradation, while reranking remains somewhat more robust. Nevertheless, \ours{} continues to achieve the strongest retrieval and end-to-end QA performance. The 10-step configuration reaches a Recall$_T$ of 0.335 and an F1$_T$ of 0.139, outperforming both dense retrieval and reranking baselines.

Interestingly, the iterative search process becomes noticeably longer in this setting. The average number of executed search steps increases from 6.244 on the original benchmark to 7.665 on the transposed version. Average runtime also increases from 3.466 hours to 5.806 hours. This suggests that when conventional table structure becomes less informative, the agent requires additional search iterations to accumulate evidence and refine hypotheses.

Overall, although performance decreases for all methods, the results indicate that \ours{} remains effective even when row-oriented entity organization is disrupted. This robustness is consistent with the design of iterative search and refinement, which relies less on a particular table layout and more on progressively discovering useful evidence through multiple retrieval rounds.

\section{Complete Case Study: Refine Behavior}
\label{app:refine-case-study-full}
This section provides the complete set of refine-behavior cases referenced in the ``Case Study: Refine Behavior'' part of the main text, drawn from the 10-step configuration on TCR-Bench~\citep{TCR-Bench}. TCR-Bench inherits its table groupings directly from Spider~\citep{Spider} and BIRD~\citep{BIRD}, where each source ``database'' is retained as a cluster of semantically related sibling tables derived from a single original table, with only one designated as the target table answerable for a given query and the rest as its sibling variants; these clusters function as groupings of table variants rather than as relational databases with enforced keys or joins. Beyond serving as this inherited grouping, all tables within and across clusters are treated as mutually independent entries in a single heterogeneous table repository, and the retrieval methods under study do not assume or exploit any cross-table relational structure.

\subsubsection{Successful Cases}

\textbf{1. Adapting to corpus terminology (Question\_G11\_Q0).}
Query: retrieving publication dates for books with identifier $\geq$ 9{,}930 and non-primary language. Initial patterns (\texttt{publication\_date}, \texttt{book\_identifier}, \texttt{language\_identifier}) did not match. After observing the shared schema across the gold table and its siblings (columns \texttt{book\_id}, \texttt{language\_id}), the model revised its patterns accordingly, achieved a perfect match score on the gold table, and correctly selected it among distractors using the numeric evidence (book identifier threshold). QA was answered correctly. This case is representative of queries whose semantic intent is clear but whose terminology differs from corpus-specific schema conventions.

\textbf{2. Large-scale terminology shift to avoid incorrect tables (Question\_G3\_Q0).}
Query: retrieving points-per-game for team \texttt{NYR} with bench-minor rating exactly 2.0. Initial patterns (\texttt{points\_per\_game}, \texttt{team\_id}, \texttt{bench\_minor\_rating}, etc.) retrieved soccer- and baseball-domain tables lacking the relevant column and evidence. The model then substantially revised its patterns toward abbreviated forms (\texttt{ppg}, \texttt{team\_abbr}, \texttt{bench\_rating}, \texttt{minor\_rating}), successfully retrieving the correct hockey table. This query was particularly difficult due to a large number of topically related distractor tables; both Dense and Rerank were misled by a hockey-domain distractor missing the bench-minor-rating information. The model did not make use of the \texttt{NYR} evidence in its final selection. This case illustrates terminology adaptation under substantial schema mismatch and heavy semantic interference.

\textbf{3. Shifting from schema terms to value matching (Question\_G8\_Q0).}
Query: retrieving district number for case number \texttt{JB374147} with beat number 1{,}014. Initial schema-only patterns failed to match any table. The model then recognized, during reasoning, that the case number was a highly distinctive value, and added value-based patterns (\texttt{JB374147}, \texttt{1014}, \texttt{1,014}), successfully locating the gold table. A more direct search on the table name (e.g., ``Crime'') would likely have converged faster. This case is representative of queries containing highly distinctive evidence values that can directly guide retrieval.

\textbf{4. Increasingly complex, incremental pattern refinement (Question\_G21\_Q0).}
Query: test dates with ANA level $>$ 64 and anti-cardiolipin IgM $>$ 11.5. The model progressively expanded its patterns to cover abbreviation and formatting variants (\texttt{ANA}, \texttt{AntiCardiolipin}, \texttt{IgM}, with and without hyphenation/whitespace), eventually matching the gold table's column names (\texttt{ANA}, \texttt{aCLIgM}) within the top-5 candidates. This query involved heavy use of domain-specific abbreviations; Dense and Rerank-3 failed to match, while Rerank-5 succeeded. This case highlights queries requiring multiple rounds of terminology exploration before convergence.

\textbf{5. Joint schema and value matching (Question\_G27\_Q0).}
Query: district corresponding to zip code 28804. The model added a value-based pattern (\texttt{28804}) alongside schema terms (\texttt{district}, \texttt{postal\_code}), correctly disambiguating the gold table from a distractor that matched on schema terms alone but corresponded to a different domain (movie-related data). This case illustrates a hybrid retrieval strategy combining schema cues with distinctive evidence values.

\textbf{6. Multi-round convergence (Question\_G33\_Q0).}
Query: average value of the foreign-only flag where the online-only flag is greater than zero. The model initially assumed underscore-separated naming conventions (e.g., \texttt{foreign\_only\_flag}) and iterated through five rounds of pattern revision, including boolean-prefix variants (\texttt{is\_foreign}, \texttt{is\_online}) and shortened forms (\texttt{foreign\_flag}, \texttt{online\_flag}), before eventually matching the gold table's camelCase columns (\texttt{isForeignOnly}, \texttt{isOnlineOnly}). The underlying query was not inherently difficult, but the incorrect initial naming assumption required substantial iteration to correct. This case demonstrates successful recovery from an initially incorrect schema assumption.

\textbf{7. Loosening overly strict patterns (Question\_G50\_Q0).}
Query: actor IDs for character name ``Maria Portokalos.'' The model's initial regex-style patterns (e.g., \texttt{actor\textbackslash s*?id}) were overly strict; relaxing them to wildcard-based forms (\texttt{actor.*id}) without introducing new terminology was sufficient to match the gold table (columns \texttt{ActorID}, \texttt{Character Name}). This case shows that refinement can improve retrieval even when the underlying terminology is already correct.

\subsubsection{Failure Cases}

\textbf{1. Premature commitment based on semantic relevance (Question\_G15\_Q0).}
Query: row GUID values where address type ID $\neq$ 2 and business entity ID $\neq$ 11{,}114. After generating patterns matching common schema terms (\texttt{addresstypeid}, \texttt{businessentityid}, \texttt{rowguid}), the model selected a distractor table sharing these column names by construction, without verifying it contained the evidence needed to answer the query. The gold table used differently abbreviated column names (\texttt{EntityBizID}, \texttt{AddrTypeID}, \texttt{guidRow}) within the same table group. The model had retrieved other sibling tables from the same group sharing partial schema overlap, which could have prompted further verification, but it did not use this signal to reconsider its selection. This case illustrates how semantic relevance can occasionally outweigh evidence verification when multiple highly related candidate tables are present.

\textbf{2. Exploration failure and premature abandonment (Question\_G58\_Q0).}
Query: maximum vertical position where price $>$ 1.25 and horizontal position $\geq$ 0.495. The model explored both literal and abbreviated coordinate-related patterns (\texttt{x\_coord}, \texttt{y\_coord}, \texttt{x\_position}, \texttt{y\_position}) and price-related terms, but ultimately emitted an empty action (\texttt{refine-none}) rather than continuing, despite the correct columns (\texttt{xpos}, \texttt{ypos}) being a plausible next variant. A value-based search strategy was also available but not attempted. This case represents insufficient exploration despite the existence of plausible refinement directions.

\textbf{3. Extended failure despite locating sibling tables (Question\_G64\_Q0).}
Query: opening values for the period between April 13, 1992 and September 27, 1992 in the American League. The model iterated through several pattern families (year- and league-specific terms, then progressively broader baseball-related terms) and located sibling tables early on, but continued making large pattern revisions rather than converging, ultimately exhausting its step budget. This query involved a large number of topically related sports distractors. This case highlights the difficulty of maintaining a stable refinement trajectory in the presence of many semantically similar candidates.

\textbf{4. Overly broad patterns without further refinement (Question\_G74\_Q0).}
Query: account balances where transaction amount $\leq$ 11. The model's revised patterns (\texttt{account}, \texttt{balance}, \texttt{amount}, \texttt{transaction}, \texttt{financial}) were overly generic, matching an unrelated table (\texttt{world\_development\_indicators::Footnotes}) rather than converging on more specific terms. This case demonstrates how insufficiently discriminative patterns can derail later refinement.

\textbf{5. Shallow understanding of corpus structure (Question\_G131\_Q1).}
Query: average day identifier where business ID equals 14676. The model attempted value-based matching on the literal numeral \texttt{14676}, but the underlying value was stored in spelled-out form (``fourteen thousand, six hundred and seventy-six''). The model had already retrieved several sibling tables early in the process (differing only in an index suffix), which could have suggested inspecting their value formatting before generating new patterns, but it did not use this signal, instead reverting to generic schema terms. This reflects a limitation in longer-horizon planning: the model does not anticipate that further iterations remain available and instead attempts to resolve the query within a single step. This case highlights challenges arising from alternative value representations and longer-horizon exploration requirements.

\section{Limitations}
Our work has several limitations that may be addressed in future research.
First, our experiments focus on a limited set of open-weight language models. Since iterative evidence seeking relies heavily on the reasoning capability of the underlying model, it remains an open question how the proposed framework behaves with stronger language models and future model families.

Second, the current framework is intentionally designed as a relatively simple proof-of-concept system to isolate the effect of iterative evidence seeking. We do not explore additional optimization techniques such as agent training, trajectory optimization, or retrieval-specific fine-tuning, all of which may further improve effectiveness and efficiency.

Third, existing table retrieval benchmarks mainly contain relatively small or medium-sized tables. The proposed framework is particularly motivated by large and heterogeneous tables, where relevant evidence may occupy only a small portion of the table. Additional evaluation on larger-scale and more realistic table repositories would provide a more comprehensive assessment of the proposed approach.

\section{Ethical Statement}
This work uses only publicly available benchmark datasets and does not involve human subjects, personal data collection, or user interaction.

The proposed framework is intended to improve evidence-oriented retrieval over structured data and does not generate new benchmark annotations or synthetic datasets. As with other retrieval and question-answering systems, the outputs may inherit limitations or errors present in the underlying datasets and language models.

We note that our contribution is methodological and benchmark-agnostic; therefore, application to real-world, sensitive domains (e.g., healthcare, finance) should be preceded by thorough, domain-adapted testing and the integration of appropriate safeguards.

Generative AI tools were used solely for code assistance and writing refinement. All research decisions, experiments, analyses, and conclusions were determined and verified by the authors.


\end{document}